\documentclass{article}

\usepackage{arxiv}
\usepackage{float}
\usepackage[utf8]{inputenc} 
\usepackage[T1]{fontenc}    
\usepackage{hyperref}       
\usepackage{url}            
\usepackage{booktabs}       
\usepackage{amsfonts}       
\usepackage{nicefrac}       
\usepackage{microtype}      
\usepackage{amsmath}
\usepackage{lipsum}
\usepackage{graphicx}
\usepackage{nth}
\begin{document}
\title{Susceptibility of Continual Learning Against Adversarial Attacks}

\author{ 
    Hikmat Khan \\
	Department of Computer Science\\
	COMSATS University Islamabad\\
	Islamabad, Pakistan \\
	\texttt{hikmat.khan179@gmail.com} \\
	\And
    Pir Masoom Shah \\
	Department of Computer Science\\
	Bacha Khan University\\
	Charsadda, KPK, Pakistan \\
	\texttt{pirmasoomshah@bkuc.edu.pk}\\
	\And
	Syed Farhan Alam Zaidi \\
	Department of Computer Science and Engineering\\
	Chung-Ang University\\
	Seoul, South Korea \\
	\texttt{syedfarhanalam1993@gmail.com}\\
	\And
	Saif ul Islam\\
	Department of Computer Science\\ Institute of Space Technology\\ 
	Islamabad, Pakistan\\
	\texttt{saiflu2004@gmail.com}\\
    \And
	Qasim Zia\\
	Department of Computer Science\\ Georgia State University\\ 
	Atlanta, United States\\
	\texttt{qzia1@student.gsu.edu}\\
}

\maketitle    
\begin{abstract}
Recent continual learning approaches have primarily focused on mitigating catastrophic forgetting. Nevertheless, two critical areas have remained relatively unexplored: 1) evaluating the robustness of proposed methods and 2) ensuring the security of learned tasks. This paper investigates the susceptibility of continually learned tasks, including current and previously acquired tasks, to adversarial attacks. Specifically, we have observed that any class belonging to any task can be easily targeted and misclassified as the desired target class of any other task. Such susceptibility or vulnerability of learned tasks to adversarial attacks raises profound concerns regarding data integrity and privacy. To assess the robustness of continual learning approaches, we consider continual learning approaches in all three scenarios, i.e., task-incremental learning, domain-incremental learning, and class-incremental learning. In this regard, we explore the robustness of three regularization-based methods, three replay-based approaches, and one hybrid technique that combines replay and exemplar approaches. We empirically demonstrated that in any setting of continual learning, any class, whether belonging to the current or previously learned tasks, is susceptible to misclassification. Our observations identify potential limitations of continual learning approaches against adversarial attacks and highlight that current continual learning algorithms could not be suitable for deployment in real-world settings. 


\end{abstract}

\section{Introduction}
Deep neural networks have achieved superhuman-level accuracy in various tasks, including image classification, semantic segmentation, biomedical image analysis, speech recognition, natural language processing, aviation, and playing games \cite{cai2020review,otter2020survey,jeong2022systematic,pandey2021comprehensive,yadav2020sentiment,le2021deep,khan2020cascading,Hikmat_75,Hikmat_76,Hikmat_77}. Collective progress across various scientific disciplines is imperative for the realization of the longstanding aspiration of Artificial General Intelligence (AGI) \cite{zeng2022introduction}. AGI necessitates that an artificial agent possesses two fundamental human-like characteristics, in addition to other intelligent behaviors: 1) adaptable and lifelong (or continual) learning capability, i.e., the ability to acquire new concepts and adapt to new environments without forgetting previously acquired knowledge. 2) preserved and robust memories, i.e., the ability to safeguard previously acquired skills \cite{zeng2022introduction}. Continual learning capabilities and the security of historical memory are integral components of Artificial General Intelligence (AGI). The research community has adopted two distinct approaches in response to these imperatives. Firstly, continual learning researchers actively concentrate on the development of algorithms capable of consistently acquiring new concepts without forgetting previously acquired abilities \cite{parisi2019continual,delange2021continual}. Secondly, within the field of adversarial machine learning, researchers have concurrently highlighted the vulnerabilities of standard algorithms to adversarial attacks \cite{wang2019security,chakraborty2018adversarial}.

Continual learning, as a critical component of machine learning paradigms, has seen considerable advancement in recent years. However, this study seeks to shed light on the vulnerabilities inherent in existing continual learning approaches when confronted with adversarial attacks. The observed susceptibility of these approaches to such attacks poses a significant challenge and necessitates a comprehensive examination. This paper outlines the empirical findings regarding these vulnerabilities and advocates for a balanced approach that addresses both the advancement of continual learning methodologies and the mitigation of their vulnerability to adversarial threats.

\textbf{Vulnerabilities in Existing Continual Learning Approaches}
Our empirical investigation uncovered notable vulnerabilities in existing continual learning approaches under adversarial conditions. Specifically, these vulnerabilities manifest in the form of misclassifications, wherein any class, irrespective of whether it belongs to the current or previously learned tasks, can be easily misclassified into the desired task class. This alarming susceptibility calls into question the reliability and robustness of these approaches in real-world scenarios.

\textbf{Misclassification of Task Classes:} One of the prominent vulnerabilities observed is the misclassification of task classes. Adversarial attacks can manipulate the decision boundaries of existing continual learning models, leading to misclassifications. Consequently, the integrity of the learned knowledge is compromised, hindering the model's ability to adapt to new tasks effectively.

\textbf{Lack of Adversarial Robustness:} Another critical vulnerability is the lack of adversarial robustness in current continual learning approaches. Adversarial examples crafted to exploit vulnerabilities in the model's decision-making process can successfully deceive the model, posing a significant threat to its reliability and usability.

\textbf{Implications for the Continual Learning Research Community:} The vulnerabilities identified in existing continual learning approaches necessitate urgent attention from the continual learning research community. It is imperative that researchers not only focus on advancing the state-of-the-art continual learning methodologies but also consider the vulnerability aspect of their proposed methods. Failure to address these vulnerabilities could impede the practical applicability of continual learning techniques in critical domains, such as autonomous vehicles, healthcare, and cyber security.

This paper presents an empirical study that illuminates several limitations within existing continual learning approaches when subjected to adversarial attacks. Our observations reveal that current continual learning algorithms are susceptible to adversarial attacks, allowing any class, whether belonging to the current or previously learned tasks, to be easily misclassified into the desired task class. 
We consider three regularization-based experiments i.e., Elastic Weight Consolidation (EWC) \cite{ewc}, Elastic Weight Consolidation Online (EWC online) \cite{ewc} and Synaptic Intelligence (SI) \cite{si}), and three replay based i.e., Learning without Forgetting (Lwf) \cite{lwf}, Deep Generative Replay (DGR) \cite{dgr} and Deep Generative Replay with Distillation (DGR + Distill) \cite{dgr}) and iCarl \ref{fig:icarl_cil} .

This paper presents an empirical study that illuminates several limitations within existing continual learning approaches when subjected to adversarial attacks. Our observations reveal that current continual learning algorithms are susceptible to adversarial attacks, allowing any class, whether belonging to the current or previously learned tasks, to be easily misclassified into the desired task class. We have undertaken a comprehensive exploration involving three regularization-based experiments, namely, Elastic Weight Consolidation (EWC) \cite{ewc}, Elastic Weight Consolidation Online (EWC online) \cite{ewc}, and Synaptic Intelligence (SI) \cite{si}, in conjunction with three replay-based methodologies, specifically, Learning without Forgetting (Lwf) \cite{lwf}, Deep Generative Replay (DGR) \cite{dgr}, and Deep Generative Replay with Distillation (DGR+Distill) \cite{dgr}, alongside iCarl as depicted in Figure \ref{fig:icarl_cil}. The vulnerability exhibited by existing continual learning approaches underscores the need for immediate attention from the continual learning research community. It necessitates a dual focus on advancing continual learning methodologies while concurrently addressing the vulnerability aspects in future proposals.

The main contributions of the research are as follows:
\begin{itemize}
    \item We contribute by identifying and highlighting vulnerabilities in existing continual learning approaches when subjected to adversarial attacks. Specifically, we have pointed out the susceptibility of these approaches to misclassification and the lack of adversarial robustness.
    \item We present empirical findings demonstrating the vulnerabilities in continual learning algorithms under adversarial conditions. This empirical evidence contributes to a better understanding of the limitations of current approaches.
    \item  We contribute by comprehensively evaluating different continual learning methodologies. We assess three regularization-based methods (Elastic Weight Consolidation, Elastic Weight Consolidation Online, Synaptic Intelligence) and three replay-based approaches (Learning without Forgetting, Deep Generative Replay, and Deep Generative Replay with Distillation) in various scenarios of continual learning.
    \item We emphasize the practical implications of these vulnerabilities by raising concerns about the reliability and robustness of continual learning approaches in real-world scenarios. This highlights the importance of addressing these issues.
    \item We contribute by calling for immediate attention from the continual learning research community to address the identified vulnerabilities. We encourage a dual focus on advancing continual learning methodologies while also considering and mitigating vulnerability in future research proposals.
\end{itemize}
In summary, we have made contributions that include the identification of vulnerabilities, presenting empirical evidence, conducting a comprehensive evaluation of methodologies, highlighting real-world concerns, and calling for further research in the field of continual learning and adversarial attacks.

\section{Related Work}
Research on continual learning is active and challenging \cite{ring1994continual,thrun1995lifelong}. It is challenging because of the catastrophic forgetting phenomenon, in which a model experiences rapid performance degradation on past tasks while learning the current task \cite{mccloskey1989catastrophic,ratcliff1990connectionist}. Section \ref{sec:Continual Learning} reviews the proposed state-of-the-art approaches to mitigate catastrophic forgetting. In addition to the challenge of mitigating catastrophic forgetting, modern deep learning methods are generally known to have weaker defenses against adversarial attacks. The approaches proposed highlight the weakness of the deep learning algorithms against adversaries\cite{wang2019security,chakraborty2018adversarial}. Section \ref{sec:Adversarial Machine Learning} briefly sheds light on the weaknesses of the deep learning algorithms against adversaries.

\subsection{Continual Learning}
\label{sec:Continual Learning}

We can categorize the extant methodologies designed to mitigate the phenomenon of catastrophic forgetting into three primary categories, as discussed in \cite{qu2021recent}.

\textbf{Regularization methods:}
In these approaches, significant alterations to the learned representation pertaining to prior tasks are effectively precluded. This is accomplished through the implementation of techniques such as regularization of the objective function or direct imposition of penalties on the model parameters. The central mechanism underlying these approaches revolves around the imposition of constraints on weight adjustments, rendering them less amenable to flexibility, as delineated by the loss function. Consequently, this enforces the stipulation that the acquisition of knowledge related to novel tasks should not substantially modify or minimally influence the proficiency of the model in addressing previous tasks. Typically, these methods are designed to gauge the salience of architectural parameters with precision. Notable examples encompass Elastic Weight Consolidation (EWC) \cite{ewc} and Synaptic Intelligence (SI) \cite{si}. Within the framework of the EWC methodology, paramount importance is accorded to parameters that occupy preeminent positions within the Fisher information matrix. In contrast, in the SI approach, the relative significance of parameters is tethered to their contributions to the loss function, such that those parameters exerting a more substantial influence on the loss are deemed more critical. In the overarching context of these methodologies, it is customary to incorporate an additional regularizer term as a requisite component. This inclusion serves to ensure the constancy of the network parameters over the course of the learning process
\cite{li2017learning,dhar2019learning,kirkpatrick2017overcoming,zenke2017continual,aljundi2018memory,saha2020gradient}.

\textbf{Dynamic architectural methods:}
In these approaches, the objective function remains invariant. However, the network capacity, denoted as the number of parameters, undergoes exponential expansion in response to novel tasks. This expansion takes various forms, including adding extra layers, nodes, or modules when introducing new tasks. The dynamic architecture typically functions by introducing new weights specific to each task and permitting adjustments solely within these task-specific weight sets. Parameter isolation techniques allocate distinct subsets of the model's parameters to each task alongside a potentially shared component
\cite{Ergn2020ContinualLW,rusu2016progressive}.

\textbf{Memory-based methods:}
In these approaches, it is observed that a portion of prior knowledge is intentionally retained for the purpose of subsequent utilization, akin to a rehearsal process, as evidenced by the scholarly works of iCaRL \cite{icarl}, Averaged Gradient Episodic Memory (A-GEM) \cite{lopez2017gradient}, Gradient Catastrophic Forgetting \cite{robins1995catastrophic}, Learning to Learn without Forgetting \cite{riemer2018learning}, and Continual Learning with Hyper networks \cite{chaudhry2020continual}. Among these approaches, the most renowned is iCaRL, a method characterized by its capacity to acquire knowledge in a class-incremental manner by preserving samples proximate to the centroids of each class within a fixed memory storage \cite{icarl}. Additionally, A-GEM represents another noteworthy example within this paradigm, as it constructs a dynamic episodic memory repository of parameter gradients during the course of the learning process \cite{lopez2017gradient}.

\subsection{Adversarial Machine Learning}
\label{sec:Adversarial Machine Learning}
An adversarial attack involves the subtle modification of an original input so that the changes are nearly invisible or practically imperceptible to the naked human eye. The modified or altered input is considered an adversary and is misclassified when presented to an original classifier, while the unmodified input remains correctly classified \cite{fgsm, pgd, cw}. The most frequently used modification measures are various Euclidean norms (e.g., $L_1$, $L_2$, $L_{\infty}$, etc.), which quantify changes at individual pixels \cite{fgsm, pgd, cw}. In real-life scenarios, adversarial attacks can be severe, compromising the data's integrity and raising questions about safety-critical applications. For instance, an autonomous vehicle may misinterpret a traffic sign, leading to an accident. The most prevalent type of adversarial attack is called an "evasion attack." In evasion attacks, an adversarial example is fed to the network, similar to its untempered counterpart, but completely confuses the classifier. It is important to note that an adversarial attack occurs during the test phase and does not modify or affect the original training data.

\textbf{Black-box Attacks vs. White-box Attacks:} Adversarial attacks can be broadly classified into black-box attacks and white-box attacks. Black-box attacks do not require access to the model's parameters \cite{fgsm}; they only require access to the model's output. Conversely, white-box attacks require full access to a model's parameters, hyper-parameters, and architecture details \cite{pgd, cw}.

\textbf{Targeted vs. Untargeted Adversarial Attacks:}
Adversarial attacks can be further classified as targeted adversarial attacks and untargeted adversarial attacks. In targeted adversarial attacks, the attacker manipulates the source input so that the classifier predicts the input as belonging to a specific target class that differs from the actual input class. In contrast, untargeted adversarial attacks aim to craftily alter the input to misclassify it as any non-target class. In other words, non-targeted attacks intend to slightly modify the source input to misclassify the perturbed input into any class except the true class. In contrast, targeted attacks aim to modify the source input to misclassify the perturbed input into the target (desired) class, except for the true class.

\textbf{One-shot vs. Iterative Adversarial Attacks:} It is worth noting that most successful attacks use gradient-based techniques, wherein the attackers alter the input in the direction of the gradient of the loss function with respect to the input. There are two main methods for carrying out such attacks: one-shot attacks, in which the attacker takes a single step in the gradient's direction, and iterative attacks, in which multiple steps are performed instead of a single step. FGSM \cite{fgsm} is a prominent example of a one-shot adversarial attack, while PGD and CW are well-known examples of iterative adversarial attacks \cite{pgd, cw}.

In this paper, we examine the vulnerability of continual learning to FGSM (i.e., the Fast Gradient Sign Method \cite{fgsm}), PGD (i.e., Projected Gradient Descent \cite{pgd}), and CW (i.e., Carlini Wagner \cite{cw}) in the context of continual learning methods, which are categorized in Section \ref{sec:Continual Learning}, within the three scenarios of continual learning, as described in Section \ref{sec: Three Scenarios of Continual Learning}.

\textbf{Fast Gradient Sign Method (FGSM)}, proposed in \cite{fgsm}, is an adversarial attack that operates in a single iteration. Mathematically, it can be described as follows:

\begin{equation}
    \left.x^{a d v}=x-\varepsilon \cdot \operatorname{sign}\left(\nabla_{x} J\left(x, y_{\text {target }}\right)\right)\right)
\end{equation}

Where $x$ represents the clean input signal, and $x^{adv}$ represents the perturbed input signal, also known as the adversarial input. $J(x, y_{target})$ represents the loss function, with $x$ input and with $y_{target}$ being the targeted label. The parameter $\varepsilon$ quantifies the degree of distortion introduced by the adversarial attack, affecting the input signal. A larger $\varepsilon$ value signifies a more impactful adversarial attack. $\varepsilon$ serves as a tunable hyper-parameter.

\textbf{Projected Gradient Descent (PGD)} is an iterative adversarial attack method commonly employed in computer security applications \cite{pgd}. Mathematically, PGD can be formalized as follows:

\begin{equation}
    P_{\mathcal{Q}}\left(\mathbf{x}_{0}\right)=\arg \min _{\mathbf{x} \in \mathcal{Q}} \frac{1}{2}\left\|\mathbf{x}-\mathbf{x}_{0}\right\|_{2}^{2}
\end{equation}


Here, $\mathbf{x}$ represents the original input signal, and $\mathbf{x_0}$ signifies the initial point within the input space. The set $\mathcal{Q}$ defines a constraint region or a ball centered around the original input $\mathbf{x}$. It is important to note that the PGD algorithm is computationally efficient when applied to problems that are straightforward to solve. However, this efficiency is contingent upon the nature of the constraint set $\mathcal{Q}$. Notably, for many non-convex sets, projecting onto them can be computationally challenging, rendering the PGD algorithm less economically viable. In cases where $\mathcal{Q}$ is a convex set, the optimization problem possesses a unique solution. Conversely, when $\mathcal{Q}$ is non-convex, the solution to $\mathcal{P}(\mathcal{Q})(\mathbf{x}_0)$ may not be unique, yielding multiple possible solutions. This property of non-convex sets introduces additional complexity to the optimization problem, which must be carefully considered when applying PGD in practice.

\textbf{Carlini Wagner} \cite{cw} is an iterative step adversarial attack. Mathematically, it can be described as follows:

\begin{equation}
    \underset{\operatorname{minimize}\left\|\boldsymbol{x}-\boldsymbol{x}_{0}\right\|^{2}+\iota_{\Omega}}{}
\end{equation}

\begin{equation}
    \iota_{\Omega}(\boldsymbol{x})=\left\{\begin{array}{l}
    0,\quad \quad  \quad \text { if } \max _{j \neq t}\left\{g_{j}(\boldsymbol{x})\right\}-g_{t}(\boldsymbol{x}) \leq 0 \\ 
    +\infty,\quad \quad otherwise
    \end{array}\right.
\end{equation}

In the above equation, $x$ represents the clean input signal, while $\iota_{\Omega}(\boldsymbol{x})$ represents regularization constraints.

\section{Three Scenarios of Continual Learning} \label{sec: Three Scenarios of Continual Learning}
Continual learning, a fundamental paradigm in machine learning, encompasses various scenarios that can be classified into three primary categories, as elucidated by \cite{three_scenarios}.
\subsection{Task Incremental Learning (Task-IL)}
Task Incremental Learning, hereafter referred to as Task-IL, is the most basic and straightforward scenario among the three. The model consistently receives explicit task identification information during the inference phase in this particular setting. This provision allows the model to incorporate task-specific components, which may manifest as distinct sub-modules within a neural network architecture, often denoted as "multi-headed." Each sub-module corresponds to a particular task and can be simultaneously trained alongside the task identification information. Notably, the output layer of such architectures adopts a "multi-headed" configuration, where each task possesses its dedicated output units. However, it is essential to emphasize that the remaining parameters of the network may potentially be shared across tasks.
\subsection{Domain Incremental Learning (Domain-IL)}
The second scenario, Domain Incremental Learning (Domain-IL), presents a distinctive challenge. In Domain-IL, the task identification remains concealed during inference, necessitating models to solve the task without inferring or being explicitly informed about the task identity. While the input distribution often exhibits variations, the network's output units, and the tasks' fundamental structural characteristics remain constant and unchanged.
\subsection{Class Incremental Learning (Class-IL)}
Class Incremental Learning (Class-IL), representing the third and most demanding scenario within continual learning, poses a formidable challenge. In Class-IL, the continual learning model is sequentially exposed to pairs of mutually exclusive classes extracted from the same data set. For instance, consider a data set comprising ten distinct classes. In the context of Class-IL, this data set undergoes division into mutually exclusive pairs, such as [0, 1], [2, 3], [4, 5], [6, 7], and [8, 9], which are subsequently sequentially presented to the model. The formidable nature of this scenario arises from the need to learn new classes while avoiding catastrophic forgetting of previously acquired knowledge, a task that underscores the complexity and significance of continual learning.
This classification of continual learning scenarios provides a foundational framework for understanding the diverse challenges and requirements that emerge in pursuing lifelong machine learning.

In this paper, our primary focus is an in-depth exploration of the robustness of continual learning methodologies when subjected to adversarial attacks. We investigate the performance of these approaches within the context of the three scenarios delineated earlier, namely Task Incremental Learning (Task-IL), Domain Incremental Learning (Domain-IL), and Class Incremental Learning (Class-IL). Our inquiry delves into how these continual learning paradigms can withstand and mitigate the disruptive influence of adversarial attacks.
To assess the resilience of continual learning methods against adversarial perturbations, we employ a spectrum of well-established adversarial attack techniques, including but not limited to the Fast Gradient Sign Method (FGSM), Projected Gradient Descent (PGD), and Carlini-Wagner (CW) attacks. These adversarial attacks, each with its distinct characteristics, are designed to craft subtle perturbations in the input data that can lead to misclassification or degradation in the performance of machine learning models. By subjecting continual learning approaches from all three scenarios to these adversarial challenges, we aim to uncover vulnerabilities and evaluate their robustness in the face of such threats.
Our preliminary findings reveal that continual learning methodologies exhibit susceptibilities to various adversarial attacks contrary to conventional expectations of robustness. These vulnerabilities represent a critical concern in real-world applications, where machine learning models' security and reliability are paramount. The implications of these findings underscore the need for comprehensive strategies to enhance the security and robustness of continual learning methods when deployed in adversarial environments.
In the subsequent section of this paper, we will delve into the methodological aspects, elucidating the precise techniques and procedures employed to craft the adversarial attacks used in our evaluations. This exploration will provide a comprehensive understanding of the mechanisms underlying these attacks and serve as a foundational framework for our subsequent discussions on the vulnerability of continual learning approaches.

\section{Methodology}

In this section, we present the methodology employed in our investigation of the vulnerabilities and susceptibility of state-of-the-art continual learning algorithms to adversarial attacks. Our study focuses on three primary aspects: the selection of continual learning algorithms, the choice of data sets for experimentation, the evaluation metrics employed, and the design of adversarial attacks. We discuss each of these aspects in detail.

Despite notable advancements in continual learning, we posit that even state-of-the-art algorithms are susceptible to both catastrophic forgetting and adversarial attacks, which can result in the misclassification of previously learned tasks. To investigate these vulnerabilities comprehensively, we selected the top-performing algorithms in various continual learning scenarios. Specifically, we considered three regularization-based algorithms: Elastic Weight Consolidation (EWC) \cite{ewc}, EWC online \cite{ewc}, and Synaptic Intelligence (SI) \cite{si}. Additionally, we evaluated three replay-based methodologies: Learning without Forgetting (Lwf) \cite{lwf}, Deep Generative Replay (DGR) \cite{dgr}, and Deep Generative Replay with Distillation (DGR+Distill) \cite{dgr}. Furthermore, we included a hybrid approach that combines replay and exemplar methods, represented as iCarl \cite{icarl}.

Regarding adversarial attacks, we opted to employ three prominent and widely recognized adversarial attack methods, namely, the Fast Gradient Sign Method (FGSM) \cite{fgsm}, Projected Gradient Descent (PGD) \cite{pgd}, and the Carlini-Wagner (CW) attack \cite{cw}. In our experimental investigations, we empirically demonstrated the vulnerability of each learned task within the context of continual learning methodologies to these three adversarial attacks, specifically, FGSM \cite{fgsm}, PGD \cite{pgd}, and CW \cite{cw}.

\subsection{Data Collection}
In our experimental investigations, we employed the widely recognized MNIST data set \cite{lecun2010mnist}. This data set comprises handwritten digits ranging from 0 to 9 and has established itself as a standard benchmark data set for training various continual learning algorithms \cite{three_scenarios}. The MNIST data set played a pivotal role in our evaluation across different continual learning scenarios. Specifically, we utilized Split MNIST for training continual learning models in the context of task incremental learning, while permuted MNIST was employed in the domain incremental learning setting. Furthermore, the unaltered MNIST data set was harnessed to train continual learning algorithms in class incremental learning scenarios, where classes (digits) were introduced sequentially to simulate real-world learning conditions.

\subsection{Evaluation Metric}

In our experiments, the models underwent training in three scenarios of continual learning: task incremental learning, domain incremental learning, and continual learning. To elaborate on these scenarios, we implemented the following procedures: In the task incremental learning, we partitioned the MNIST data set into five separate tasks, with each task encompassing two distinct classes. In domain incremental learning, we divided the MNIST data set into two tasks, with each task comprising five classes from the MNIST data set. We segmented the MNIST data set into nine mutually exclusive tasks in class incremental learning. The initial task included two classes; subsequently, one class was incrementally added for each subsequent task. To evaluate the performance of our models across these scenarios, we calculated the average accuracy by aggregating the results of all experiments. The average accuracy ($ACC$) was computed using the following formula, as described in \cite{three_scenarios}:

\begin{equation} ACC = \frac{1}{T} \sum_{i=1}^{T} R_{T,i}, \end{equation}

Here, $R$ represents the average accuracy, while $i$ corresponds to the task index.

\subsection{Training Protocol}
We employed the code made publicly available by the by \cite{three_scenarios} to conduct training on all three scenarios of the continual learning methodologies, including Elastic Weight Consolidation (EWC) \cite{ewc}, Online EWC \cite{ewc}, Synaptic Intelligence (SI) \cite{si}, XDG \cite{xgd}, Learning without Forgetting (LwF) \cite{lwf}, Deep Generative Replay (DGR) \cite{dgr}, DGR with knowledge distillation (DGR + distill), and Incremental Classifier and Representation Learning (ICARL) \cite{icarl}. We conducted the training on the MNIST data set \cite{lecun2010mnist}. It is worth noting that all hyper-parameter configurations remained consistent with the original specifications outlined in \cite{three_scenarios}. Consequently, implementing these continual learning approaches yielded results aligned with the standard evaluation accuracy levels, as originally reported in  \cite{three_scenarios}. Each experiment was repeated 20 times to ensure the robustness of our findings, each with a distinct random seed. This approach was adopted to mitigate potential sources of variability and to provide a more accurate estimation of the average accuracy across multiple runs. Subsequently, the mean average accuracy and the corresponding standard deviations were computed to capture the variability inherent in the results across these repeated experiments.

\subsection{Designing an Adversarial Attacks}
This paper investigates the resilience of continual learning algorithms in the face of adversarial attacks. Precisely, we assess the performance of these algorithms when subjected to three standard adversarial attack methods, namely Fast Gradient Sign Method (FGSM), Projected Gradient Descent (PGD), and Carlini-Wagner (CW) attacks. We employ the open-source Python tool "foolbox" to conduct these experiments to generate adversarial attacks, following the methodology proposed by \cite{rauber2017foolboxnative}.
Our study explores the reliability of the algorithms in three distinct continual learning scenarios, considering both targeted and untargeted attacks. Targeted attacks represent the most potent form of adversarial attacks, while untargeted attacks are comparatively less powerful. The untargeted attacks are a more efficient but often less accurate method of executing targeted attacks, wherein the attacker aims to misclassify the input into any class closest to the desired target.
Our empirical findings reveal that continual learning algorithms are highly susceptible to adversarial attacks in all scenarios examined. In summary, we demonstrate that any learned task, whether it pertains to current or historical learning, can be abruptly attacked and misclassified into a class desired by the adversary. Intriguingly, our investigation uncovers that tasks learned in the past exhibit a higher vulnerability to misclassification than those learned more recently. This heightened susceptibility to misclassification gives rise to the creation of false memories within artificial agents. Such false memory formation significantly hinders the deployment of artificial agents in real-world applications, particularly in safety-critical domains like autonomous vehicles.

\section{Results and Discussion}
We examined the robustness of continual learning methods concerning adversarial attacks in a broader context, particularly emphasizing the security of individually learned tasks. Our investigation reveals that adversarial attacks compromise state-of-the-art continual learning models. Specifically, we demonstrate that any learned task within the continual learning paradigm can be susceptible to adversarial attacks, resulting in misclassification.
A noteworthy observation from our study is that newly acquired tasks exhibit a lower vulnerability to adversarial attacks than previously learned tasks. This observation brings to light the intriguing phenomenon wherein creating false memories associated with historically learned tasks appears to be a more tractable endeavor in the context of adversarial attacks.
\begin{table*}[h]

    \centering
 \caption{The \nth{1} and \nth{2} columns present the results of continual learning approaches and their corresponding accuracy scores, which are averaged across tasks and obtained under standard evaluation settings in task incremental scenarios. The \nth{3}, \nth{4}, and \nth{5} columns display the reductions in average accuracy resulting from FGSM, PGD, and CW adversarial attacks, respectively. The labels "U" and "T" in columns three through five denote untargeted and targeted adversarial attacks. The decrease in average accuracy, as depicted in columns \nth{3} through \nth{5}, underscores the efficacy of adversarial attacks and illustrates the susceptibility of any learned task to successful misclassification. It is important to note that each experiment was independently replicated 20 times using different random seeds to ensure a more robust approximation of the results.}   
    \begin{tabular}{
   |p{2.25cm}|p{2cm}|p{2.7cm}|p{2.7cm}|p{2.7cm}|
    }
    
     \hline
     \multicolumn{5}{|c|}{Task IL Setting \cite{three_scenarios}}  \\
     \hline
      \hline
        \textbf{Approach} & \textbf{Task-IL}  & \textbf{FGSM} \cite{fgsm} &  \textbf{PGD} \cite{pgd} & \textbf{CW} \cite{cw} \\
         \hline
        EWC \cite{ewc} & $98.5\% (\pm0.7)$ & $23.6\% (\pm10.5)$-U \newline$32.0\%(\pm10.71)$-T&   $ 63.1\%(\pm6.5)$-U\newline $83.8\%(\pm6.32)$-T&  $98.5\%(\pm0.7)$-U\newline  $77.6\%(\pm9.38)$-T  \\
         \hline
        EWC Online \cite{ewc} & $98.2\%(\pm1.4)$ & $18.9\%(\pm8.0)$-U\newline $33.2\%(\pm10.6)$-T & $59.3\%(\pm8.2)$-U \newline $83.0\%(\pm6.69)$-T  & $98.6\%(\pm0.7)$-U\newline$84.4\%(\pm6.27)$-T   \\
        \hline 
        SI \cite{si} & $87.8\%(\pm7.5)$ & $23.2\%(\pm6.3)$-U \newline $35.8\%(\pm9.9)$-T& $62.8\%(\pm7.0)$-U \newline $78.7\%(\pm7.99)$-T & $87.8\%(\pm7.5)$-U \newline $68.8\%(\pm10.22)$-T     \\
        \hline 
        XDG \cite{xgd} & $84.7\%(\pm7.5)$ & $32.3\%(\pm5.1)$-U\newline $36.9\%(\pm10.01)$-T & $84.7\%(\pm7.5)$-U \newline $66.6\%(\pm9.31)$-T & $84.7\%(\pm7.5)$-U \newline $66.6\%(\pm9.31)$-T    \\
        \hline 
        
        LwF \cite{lwf} & $99.4\%(\pm0.2)$ & $14.0\%(\pm8.1)$-U\newline $34.8\%(\pm8.12)$-T& $70.5\%(\pm6.7)$-U \newline $91.6\%(\pm4.29)$-T  &  $99.4\%(\pm0.2)$-U \newline  $83.4\%(\pm5.58)$-T      \\
        \hline 
        DGR \cite{dgr} & $99.5\%(\pm0.2)$ & $53.3\%(\pm9.8)$-U\newline $34.2\%(\pm8.17)$-T & $81.1\%(\pm8.1)$-U \newline $92.3\%(\pm4.27)$-T & $99.5\%(\pm0.2)$-U \newline $82.9\%(\pm4.88)$-T    \\
        \hline 
        DGR + Distill \cite{dgr} &$99.5\%(\pm0.2)$ & $22.5\%(\pm6.1)$-U\newline $31.0\%(\pm8.28)$-T & $65.1\%(\pm6.0)$-U \newline $90.5\%(\pm4.88)$-T  & $99.5\%(\pm0.2)$-U \newline $84.2\%(\pm4.91)$-T   \\
        \hline

     \end{tabular}
    
    \label{table:task_il}
\end{table*}

\begin{table*}[h]

    \centering
    \caption{ The \nth{1} and \nth{2} columns of the table present the continual learning methodologies employed, as well as the corresponding accuracy scores (averaged across multiple tasks). These measurements were obtained under standard evaluation conditions within a domain incremental setting. The columns \nth{3} and \nth{4}, on the other hand, illustrate the reduction in average accuracy scores when subjected to FGSM (Fast Gradient Sign Method) and PGD (Projected Gradient Descent) adversarial attacks, respectively. Notably, the labels "U" and "T" in columns \nth{3}, \nth{4}, and \nth{5} are used to differentiate between untargeted and targeted adversarial attacks. The decline in average accuracy, specifically in columns \nth{3} and \nth{4}, underscores the effectiveness of these adversarial attacks. Furthermore, it emphasizes the vulnerability of any learned task to successful attacks, resulting in misclassification. It is important to note that each experiment was independently repeated 20 times, utilizing different random seeds for each iteration, to accurately approximate the results.}    
    \begin{tabular}{
    |p{2.7cm}|p{2.7cm}|p{2.7cm}|p{2.7cm}|
    }
    
     \hline
     \multicolumn{4}{|c|}{Domain IL Setting \cite{three_scenarios}} \\
     \hline
      \hline
        \textbf{Approach} & \textbf{Domain-IL} & \textbf{FGSM}\cite{fgsm} &  \textbf{PGD}\cite{pgd}  \\
         \hline
        EWC \cite{ewc} & $78.6\%(\pm4.8)$ &$0.0\%(\pm0.0)$-U \newline$19.1\%(\pm2.77)$-T&   $4.4\%(\pm1.7)$-U\newline $34.4\%(\pm10.14)$-T  \\
         \hline
        EWC Online \cite{ewc} & $78.2\%(\pm5.0)$ & $0.0\%(\pm0.0)$-U\newline $19.1\%(\pm2.7)$-T & $4.7\%(\pm1.7)$-U \newline $34.8\%(\pm10.22)$-T      \\
        \hline 
        SI \cite{si} & $66.3\%(\pm4.7)$ & $0.1\%(\pm0.1)$-U \newline $18.9\%(\pm2.67)$-T & $16.4\%(\pm3.0)$-U \newline $46.4\%(\pm9.02)$-T    \\
        \hline 
        XDG \cite{xgd} & $67.0\%(\pm5.2)$ & $0.3\%(\pm0.4)$-U\newline $18.8\%(\pm2.68)$-T & $11.5\%(\pm3.7)$-U \newline $40.4\%(\pm9.07)$-T    \\
        \hline 
        
        LwF \cite{lwf} & $73.6\%(\pm4.9)$ & $0.0\%(\pm0.0)$-U\newline $18.3\%(\pm2.77)$-T& $3.0\%(\pm1.7)$ \newline $32.6\%(\pm8.39)$-T     \\
        \hline 
        DGR \cite{dgr} & $96.3\%(\pm0.7)$ & $1.0\%(\pm1.1)$-U\newline $13.2\%(\pm4.8)$-T & $18.8\%(\pm2.9)$-U \newline $30.7\%(\pm9.98)$-T    \\
        \hline 
        DGR + Distill \cite{dgr} &$96.4\%(\pm0.6)$ & $0.9\%(\pm1.2)$-U\newline $14.0\%(\pm4.58)$-T &$9.9\%(\pm1.5)$-U \newline $29.2\%(\pm9.12)$-T     \\
        \hline 

     \end{tabular}

    \label{table:domain_il}
\end{table*}

\begin{table*}[h]
    \centering
    \caption{ The \nth{1} and \nth{2} columns of the table present the results of the continual learning approach, including the accuracy achieved (averaged across tasks) under standard evaluation conditions in class incremental settings. The columns \nth{3}, \nth{4}, and \nth{5} represent the decrease in average accuracies when subjected to FGSM, PGD, and CW adversarial attacks, respectively. The labels "U" and "T" in columns \nth{3}, \nth{4}, and \nth{5} denote untargeted and targeted adversarial attacks, respectively. The decline in average accuracy (i.e., in columns \nth{3}, \nth{4}, and \nth{5}) serves as an indicator of the success of these adversarial attacks and underscores the vulnerability of any learned task to successful misclassification. Each experiment was conducted 20 times with varying seed values to obtain a more robust approximation.}    
    \begin{tabular}{
    |p{2.2cm}|p{2cm}|p{2.5cm}|p{2.5cm}|p{2.5cm}|
    }
    
     \hline
     \multicolumn{5}{|c|}{Class IL Setting \cite{three_scenarios}}  \\
     \hline
      \hline
        \textbf{Approach} & \textbf{Class-IL} & \textbf{FGSM}\cite{fgsm} &  \textbf{PGD}\cite{pgd} & \textbf{CW}\cite{cw} \\
        \hline
        ICARL \cite{icarl}& $90.4\%(\pm1.0)$ &  $0.5\%(\pm0.6)$-U\newline $8.8\%(\pm1.67)$-T &$3.7\%(\pm2.2)$-U \newline $27.8\%(\pm8.72)$-T   & $0.1\%(\pm0.0)$-U \newline  $10.0\%(\pm0.41)$-T    \\
        \hline 

     \end{tabular}

    \label{table:class_il}
\end{table*}

In Tables \ref{table:task_il}, \ref{table:domain_il}, and \ref{table:class_il}, we summarized the performance of the continual learning models when attacked by adversarial attacks. The individual class-level vulnerability of the continual learning algorithms under Task-IL can be seen in Figures \ref{fig:ewc_til},\ref{fig:ewc_online_til},\ref{fig:si_til},\ref{fig:lwf_til},\ref{fig:dgr_til},\ref{fig:dgr_distill_til}.

\subsection{Analyzing Class-Wise Vulnerability of EWC to Adversarial Attacks in Task-IL Continual Learning}
\begin{figure}
\centering
\includegraphics[width=1.0\textwidth]{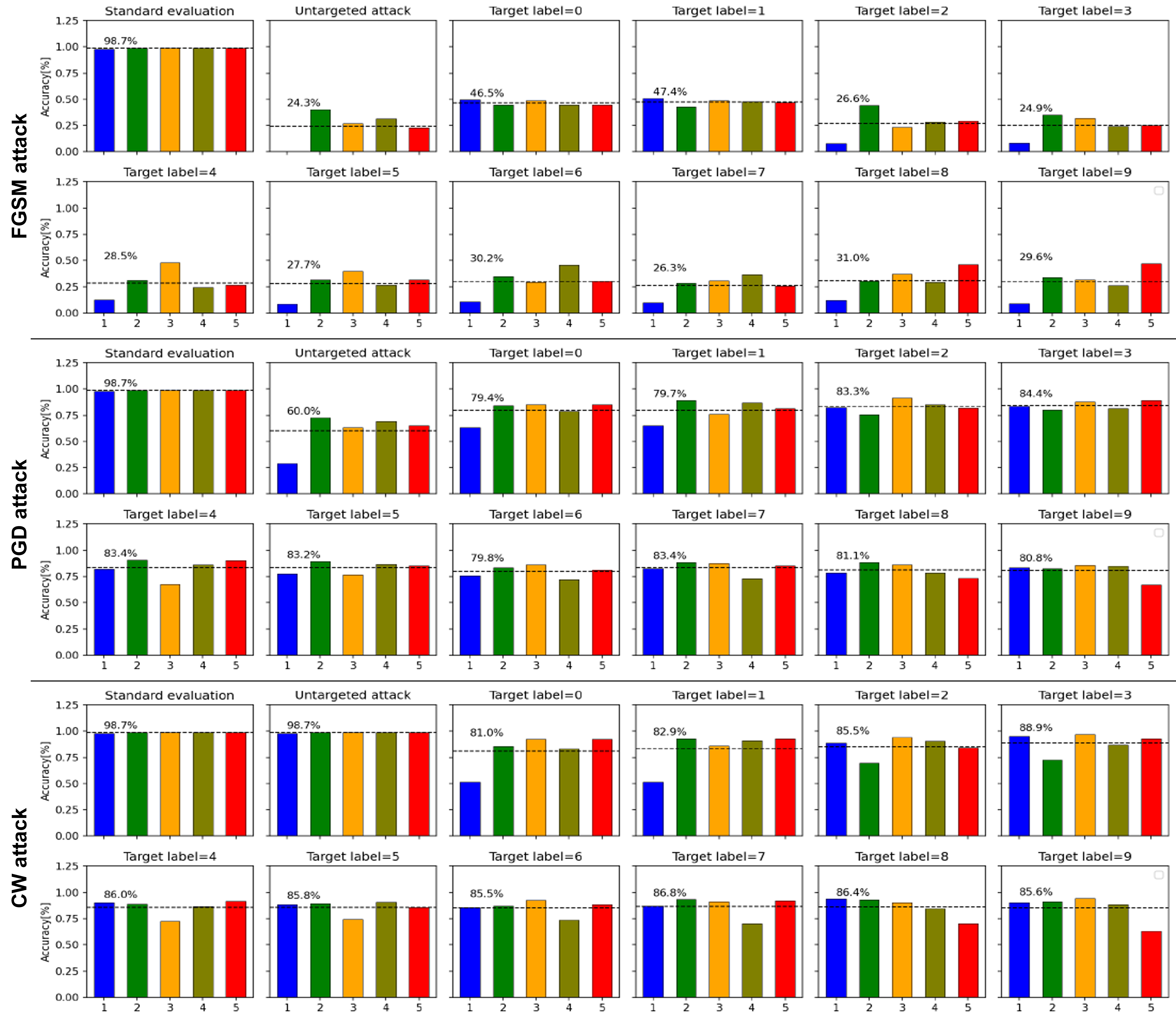}
\caption{
Class-wise vulnerability of the EWC \cite{ewc} against the FGSM, PGD, and CW \cite{fgsm,pgd,cw} adversarial attacks under Task-IL setting of continual learning.}
\label{fig:ewc_til}
\end{figure}

Figure \ref{fig:ewc_til} illustrates the class-wise vulnerability of EWC \cite{ewc} against FGSM, PGD, and CW adversarial attacks \cite{fgsm,pgd,cw} under the Task-IL setting of continual learning. The first two rows display EWC's class-wise vulnerability to FGSM attacks, while the following two rows depict its vulnerability to PGD attacks. The last two rows showcase EWC's class-wise vulnerability to CW attacks.

In addition, the first sub-figure in rows 1, 3, and 5 represents the average performance of EWC under standard evaluation conditions for continual learning. The second sub-plot in each row demonstrates the performance degradation under untargeted adversarial attacks. Subsequent sub-plots reveal how targeted adversarial attacks affect the average performance, with the sub-plot headers indicating the targeted labels.

Furthermore, the x-axis of the plots corresponds to the task number, while the y-axis represents the average accuracy over 10 runs.

\subsection{Analyzing Class-Wise Vulnerability of EWC-online to Adversarial Attacks in Task-IL Continual Learning}
\begin{figure}
\centering
\includegraphics[width=1.0\textwidth]{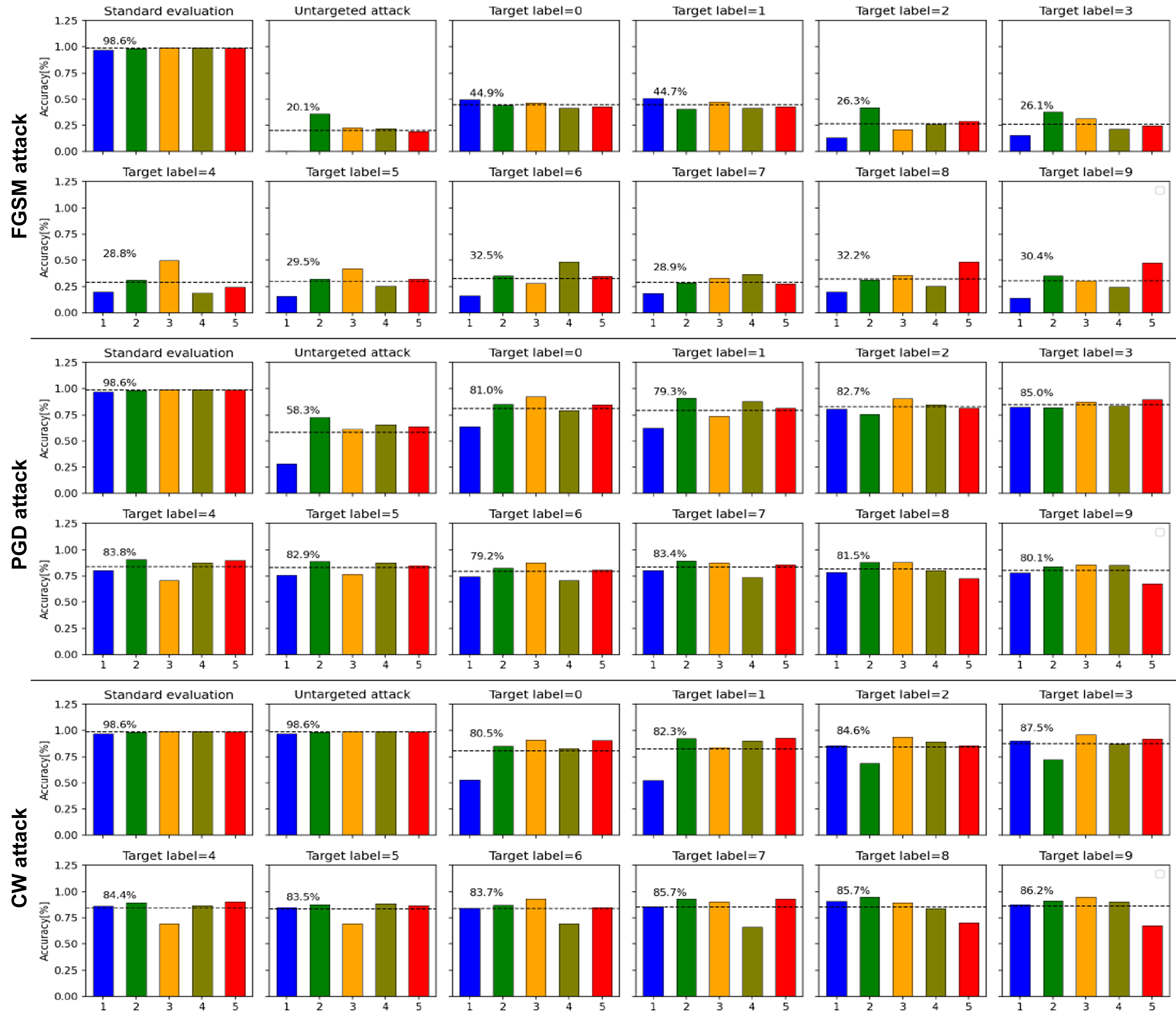}
\caption{class-wise vulnerability of the EWC online \cite{ewc} against the FGSM, PGD, and CW \cite{fgsm,pgd,cw} adversarial attacks under Task-IL setting of continual learning.}
\label{fig:ewc_online_til}
\end{figure}

In Figure \ref{fig:ewc_online_til}, we illustrate the class-wise vulnerability of the EWC online model \cite{ewc} when subjected to adversarial attacks, namely the Fast Gradient Sign Method (FGSM), Projected Gradient Descent (PGD), and Carlini-Wagner (CW) attacks \cite{fgsm,pgd,cw}, within the context of Task-Incremental Learning (Task-IL). 
The initial two rows of Figure \ref{fig:ewc_online_til} provide insights into the class-wise vulnerability of the EWC online model against FGSM attacks. In contrast, the subsequent two rows, the third and fourth, focus on the model's susceptibility to PGD attacks, and the final two rows depict the class-wise vulnerability under CW attacks.
Additionally, each row contains two sub-figures. The first sub-figure within rows 1, 3, and 5 represents the EWC online model's average performance when evaluated under standard conditions for continual learning. Conversely, the second sub-figures in these rows show the extent of degradation in model performance when subjected to untargeted adversarial attacks.
Furthermore, we explore the impact of targeted adversarial attacks on overall performance in the subsequent sub-plots. These sub-plot headers provide information regarding the specific target labels for each attack. The x-axis in all sub-plots indicates the task number, while the y-axis displays the average accuracy over ten separate runs of the experiment.

\subsection{Analyzing Class-Wise Vulnerability of SI to Adversarial Attacks in Task-IL Continual Learning}
\begin{figure}
\centering
\includegraphics[width=1.0\textwidth]{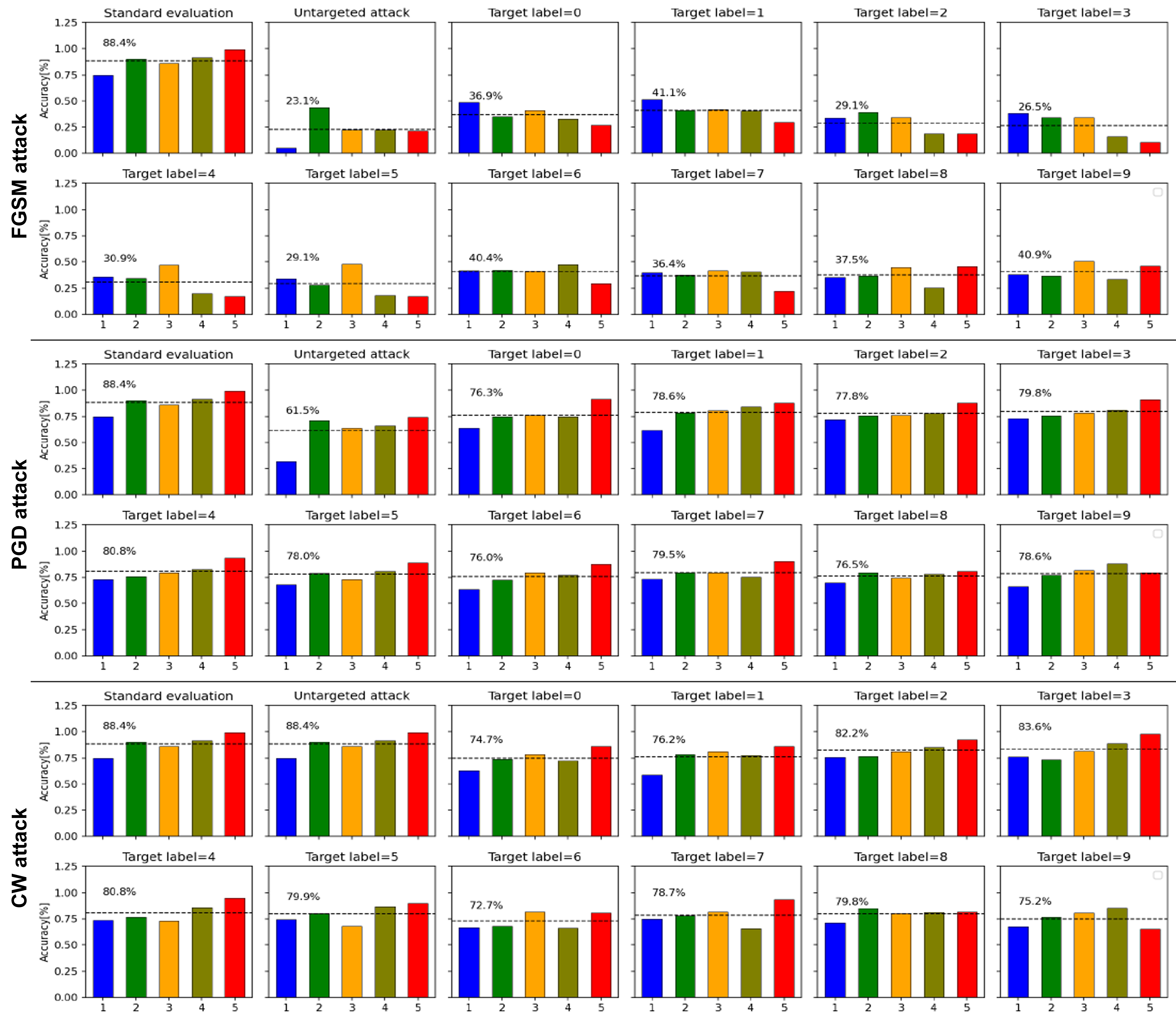}
\caption{Class-wise vulnerability of the SI  \cite{si} against the FGSM, PGD, and CW \cite{fgsm,pgd,cw} adversarial attacks under Task-IL setting of continual learning.}
\label{fig:si_til}
\end{figure}

The class-wise vulnerability of the SI \cite{si} against FGSM, PGD, and CW adversarial attacks \cite{fgsm,pgd,cw} under the Task-IL setting of continual learning is depicted in Figure \ref{fig:si_til}. The first and second rows depict the class-wise vulnerability of the SI against FGSM. Similarly, the third and fourth rows show the class-wise vulnerability of the SI against PGD, and the fifth and sixth rows present the class-wise vulnerability of the SI against CW attacks. The first sub-figure in rows 1, 3, and 5 presents the SI's average performance under standard evaluation in the context of continual learning. The degradation under untargeted adversarial attacks is depicted in the second sub-plots in rows 1, 3, and 5. Furthermore, the following sub-plots illustrate the decline in average performance due to targeted adversarial attacks, with the headers of the sub-plots indicating the targeted labels. The x-axis represents the task number, while the y-axis displays the average accuracy over ten runs.

\subsection{Analyzing Class-Wise Vulnerability of XDG to Adversarial Attacks in Task-IL Continual Learning}
\begin{figure}
\centering
\includegraphics[width=1.0\textwidth]{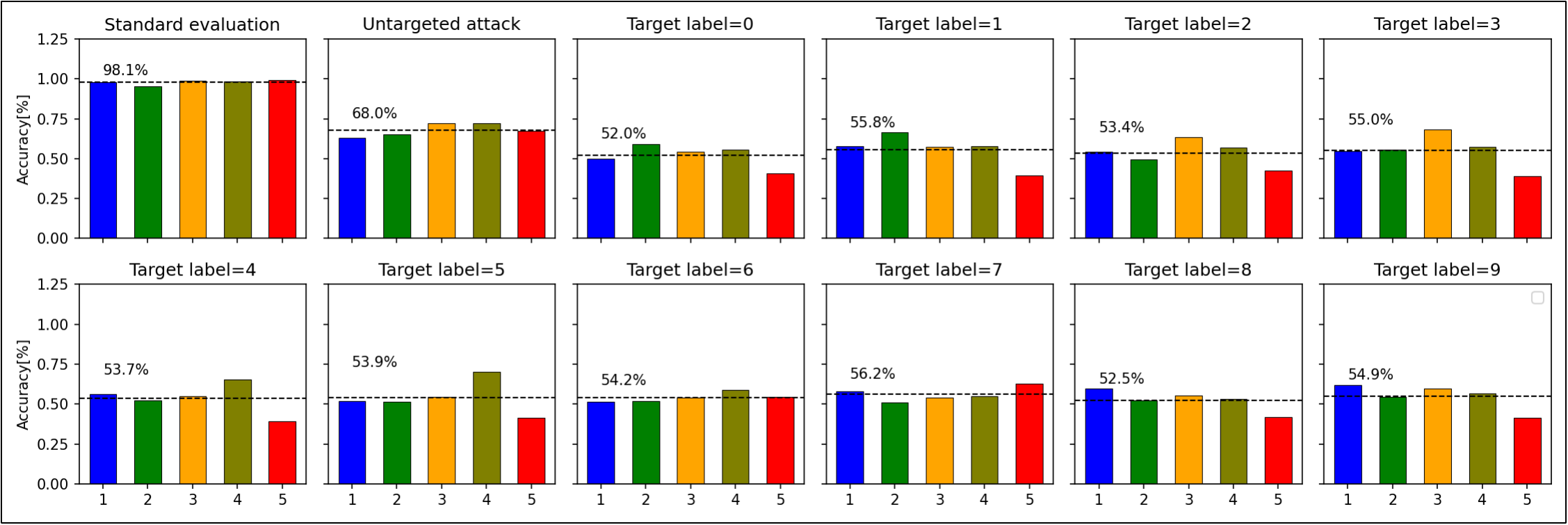}
\caption{Class-wise vulnerability of the XDG  \cite{xgd} against the FGSM \cite{fgsm} adversarial attacks under Task-IL setting of continual learning.}
\label{fig:XDG_FGSM}
\end{figure}

In Figure \ref{fig:XDG_FGSM}, we illustrate the class-wise vulnerability of the XDG model \cite{xgd} in the context of adversarial attacks using the FGSM method \cite{fgsm} under the Task-IL (Task-Incremental Learning) setting in continual learning. The targeted labels for the attacks are explicitly denoted in the headers of the subplots. The x-axis in the figure corresponds to the task number, while the y-axis represents the average accuracy obtained over ten separate runs.

\subsection{Analyzing Class-Wise Vulnerability of LwF to Adversarial Attacks in Task-IL Continual Learning}
\begin{figure}
\centering

\includegraphics[width=1.0\textwidth]{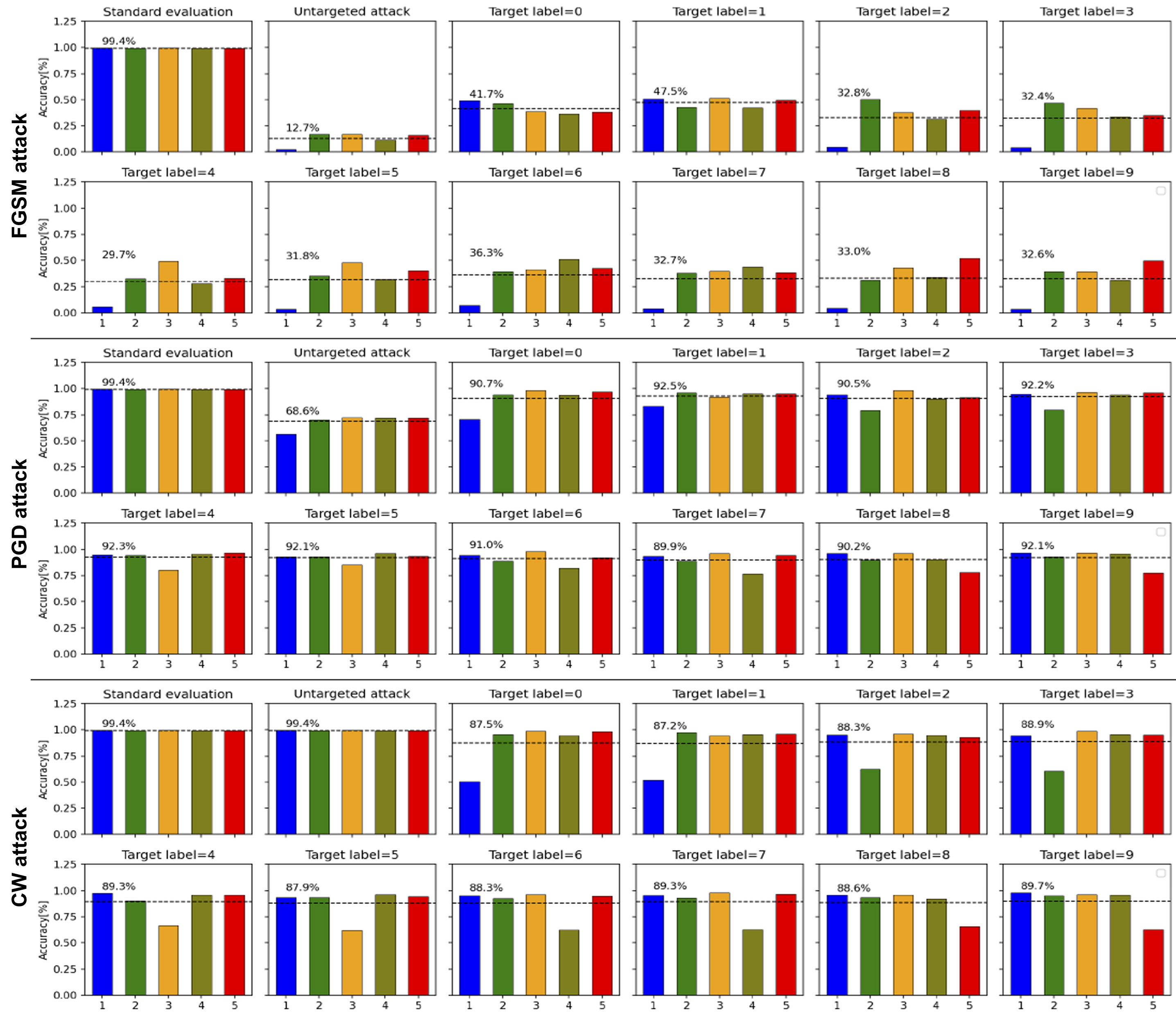}

\caption{Presents class-wise vulnerability of the Lwf  \cite{lwf} against the FGSM, PGD, and CW \cite{fgsm,pgd,cw} adversarial attacks under Task-IL setting of continual learning.}
\label{fig:lwf_til}
\end{figure}

In addition, as illustrated in Figure \ref{fig:lwf_til}, an analysis of the class-wise vulnerability of the Learning without Forgetting (Lwf) model \cite{lwf} against various adversarial attacks, including the Fast Gradient Sign Method (FGSM), Projected Gradient Descent (PGD), and Carlini-Wagner (CW) attacks \cite{fgsm,pgd,cw}, is presented within the context of the Task-IL setting in continual learning. The figure is divided into three sections, each corresponding to one type of attack.
The initial two rows of Figure \ref{fig:lwf_til} represent the class-wise vulnerability of the Lwf model when subjected to FGSM attacks. Subsequently, the following two rows depict the Lwf model's response to PGD attacks, and the final two rows showcase its vulnerability against CW attacks.
Within each set of rows (i.e., rows 1 and 2, 3 and 4, 5 and 6), the first sub-figure demonstrates the Lwf model's average performance under standard evaluation conditions for continual learning. Following this, the second sub-plot in each pair of rows illustrates the degradation in performance when the model is exposed to untargeted adversarial attacks. These plots provide insights into how the model's accuracy is affected by such attacks.
The subsequent sub-plots in each row reveal the deterioration in average performance resulting from targeted adversarial attacks, with the sub-plot headings specifying the targeted labels. This analysis sheds light on the model's robustness against specific adversarial goals.
The x-axis in all sub-plots represents the task number, while the y-axis quantifies the average accuracy over ten independent runs, offering a comprehensive view of the Lwf model's performance in the face of adversarial challenges.
In summary, Figure \ref{fig:lwf_til} provides a detailed examination of the Lwf model's performance and vulnerability under different adversarial attack scenarios in the Task-IL setting of continual learning.

\subsection{Analyzing Class-Wise Vulnerability of DGR to Adversarial Attacks in Task-IL Continual Learning}
\begin{figure}
\centering
\includegraphics[width=1.0\textwidth]{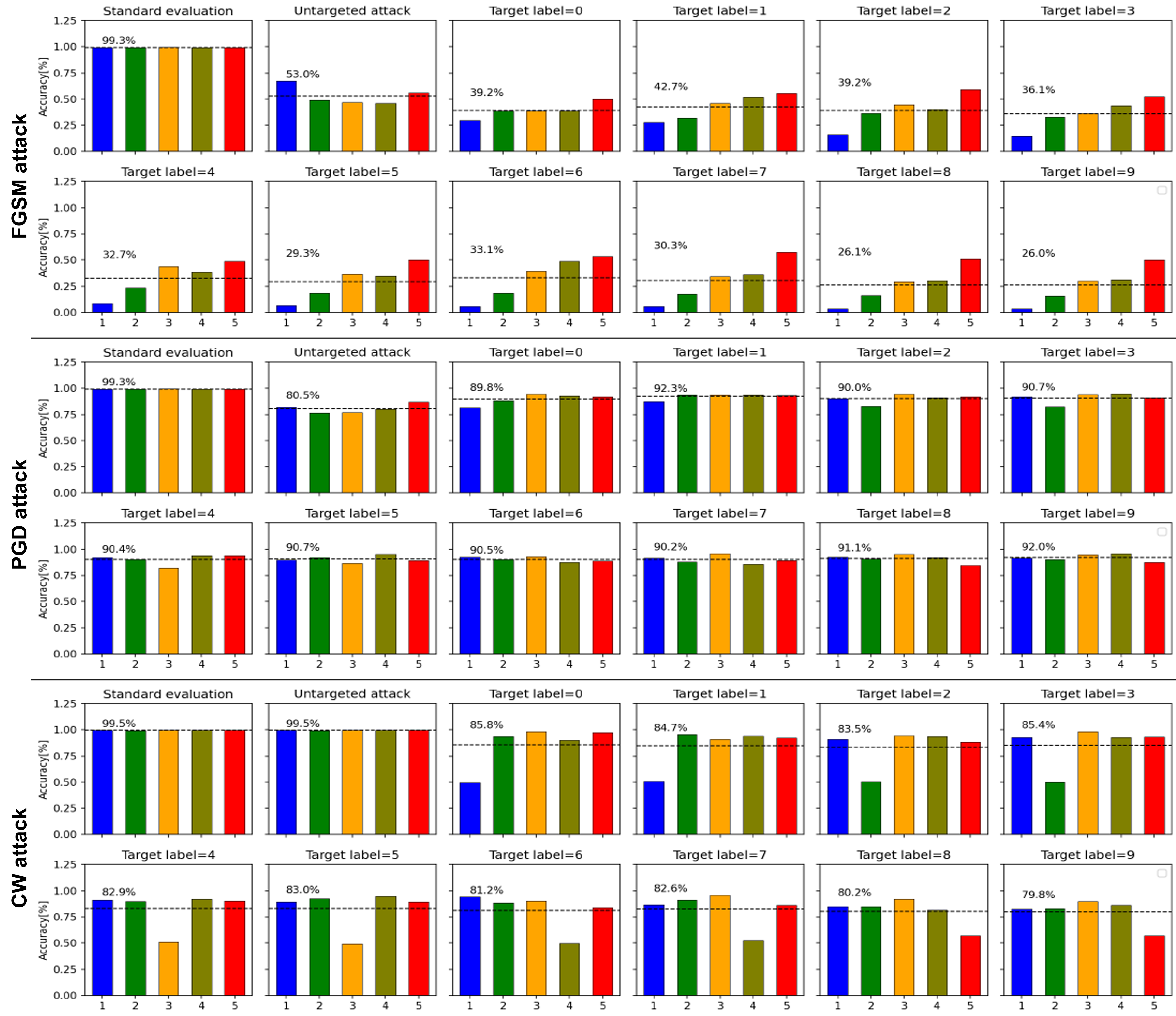}
\caption{Presents class-wise vulnerability of the DGR  \cite{dgr} against the FGSM, PGD, and CW \cite{fgsm,pgd,cw} adversarial attacks under Task-IL setting of continual learning.}
\label{fig:dgr_til}
\end{figure}

In Figure \ref{fig:dgr_til}, we present an analysis of the class-wise vulnerability of the Dynamic Group Regularization (DGR) model \cite{dgr} when subjected to various adversarial attacks, namely the Fast Gradient Sign Method (FGSM), Projected Gradient Descent (PGD), and Carlini-Wagner (CW) attacks \cite{fgsm,pgd,cw}. This investigation is conducted within the context of a Task-IL (Task-Incremental Learning) setting.
The figure is organized into rows, each corresponding to one of the three attack methods (FGSM, PGD, and CW). Within each row are two pairs of sub-figures, each representing a different aspect of the model's vulnerability.
The first pair of sub-figures (in rows 1, 3, and 5) portrays the DGR model's performance under standard evaluation conditions during continual learning. Specifically, it illustrates the average performance using the Learning without Forgetting (LwF) method.
Moving to the second pair of sub-figures (in rows 1, 3, and 5), we explore the model's degradation when exposed to untargeted adversarial attacks. These sub-figures provide insight into how the DGR model's average performance is affected when subjected to such attacks.
Finally, the last pair of sub-figures (in rows 1, 3, and 5) delves into the model's susceptibility to targeted adversarial attacks. These sub-figures draw attention to the decline in average performance under these attacks and are labeled with the targeted labels to clarify the nature of the attacks.
The x-axis of all sub-figures represents the task number, reflecting the progression of learning tasks in the continual learning setting. Meanwhile, the y-axis shows the average accuracy, calculated over ten independent runs, providing a robust assessment of the DGR model's performance in the face of adversarial challenges.

\subsection{Analyzing Class-Wise Vulnerability of DGR+Distill to Adversarial Attacks in Task-IL Continual Learning}
\begin{figure}
\centering
\includegraphics[width=1.0\textwidth]{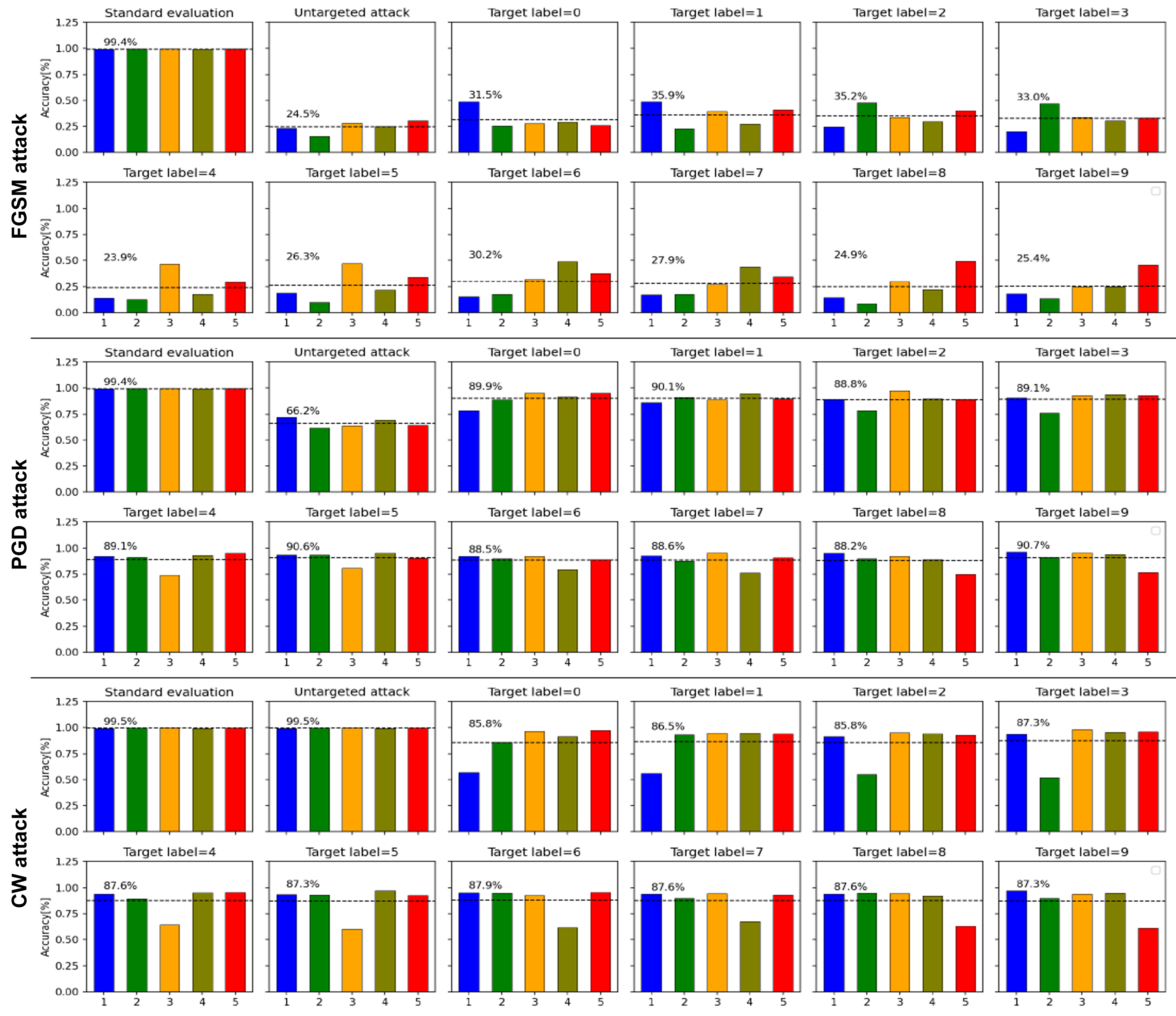}
\caption{Presents class-wise vulnerability of the DGR+Distill  \cite{dgr} against the FGSM, PGD, and CW \cite{fgsm,pgd,cw} adversarial attacks under Task-IL setting of continual learning.}
\label{fig:dgr_distill_til}
\end{figure}

Furthermore, Figure \ref{fig:dgr_distill_til} illustrates the class-wise vulnerability of the DGR+Distill model \cite{dgr} when subjected to the FGSM, PGD, and CW adversarial attacks \cite{fgsm,pgd,cw} under the Task-IL setting of continual learning. The top two rows present the class-wise vulnerability of the DGR+Distill model against FGSM attacks. Subsequently, the following two rows depict its vulnerability against PGD attacks, while the bottom two rows illustrate its vulnerability against CW attacks.
In each set of rows (1, 3, and 5), the first sub-figure showcases the average performance of the Learning without Forgetting (LwF) strategy under standard evaluation conditions in the context of continual learning. The second sub-plots in rows 1, 3, and 5 display the extent of performance degradation resulting from untargeted adversarial attacks. The subsequent sub-plots in each row provide insights into the impact of targeted adversarial attacks on the average performance of the DGR+Distill model. The headers of these sub-plots specify the targeted labels during the attacks.
Moreover, the x-axis of the plots represents the task number, while the y-axis represents the average accuracy computed over ten independent runs.

\subsection{Analyzing Class-Wise Vulnerability of EWC to Adversarial Attacks in Domain-IL Continual Learning}
\begin{figure}[H]
\centering
\includegraphics[width=1.0\textwidth]{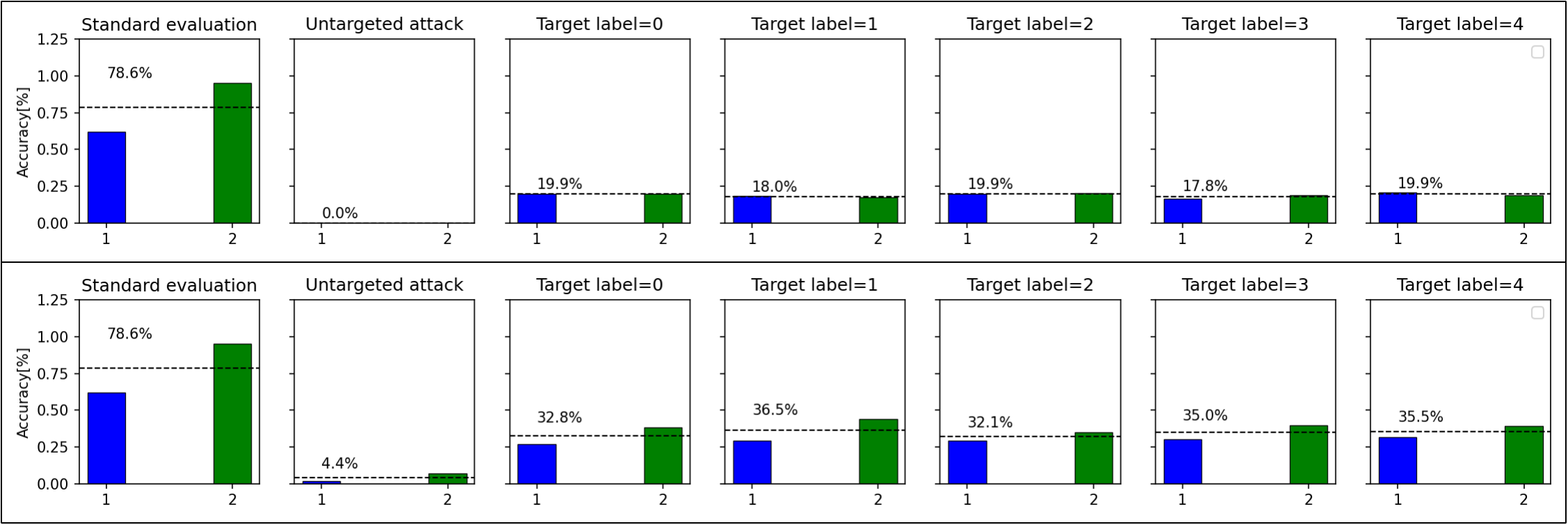}
\caption{Class-wise vulnerability of the EWC \cite{ewc} against the FGSM and PGD \cite{fgsm,pgd} adversarial attacks under Domain-IL setting of continual learning.}
\label{fig:ewc_dil}
\end{figure}

Figure \ref{fig:ewc_dil} presents the class-wise vulnerability analysis of the Elastic Weight Consolidation (EWC) model \cite{ewc} when subjected to FGSM adversarial attacks \cite{fgsm}. This investigation is conducted within the Domain-IL setting of continual learning.
The first row of the figure illustrates the class-wise vulnerability of the EWC model to FGSM attacks, while the second row showcases the class-wise vulnerability of the Incremental Classifier and Representation Learning (ICARL) model against PGD attacks.
In both rows (1 and 2), the first sub-figure provides an overview of the average performance of the EWC model when evaluated under standard conditions for continual learning. Subsequently, the second sub-plots in rows 1 and 2 visualize the extent of performance degradation resulting from untargeted adversarial attacks, specifically FGSM attacks in the case of EWC and PGD attacks for ICARL.
The subsequent sub-plots in both rows offer insights into how targeted adversarial attacks affect the overall performance of the models. The headers accompanying these sub-plots specify the labels that were the focus of the targeted attacks.
It is worth noting that the x-axis in these plots represents the task number, while the y-axis represents the average accuracy computed over ten independent runs. The horizontal bar on the plots signifies the average accuracy obtained across two consecutive tasks, providing a reference point for performance comparison.

\subsection{Analyzing Class-Wise Vulnerability of EWC-online to Adversarial Attacks in Domain-IL Continual Learning}
\begin{figure}[H]
\centering
\includegraphics[width=1.0\textwidth]{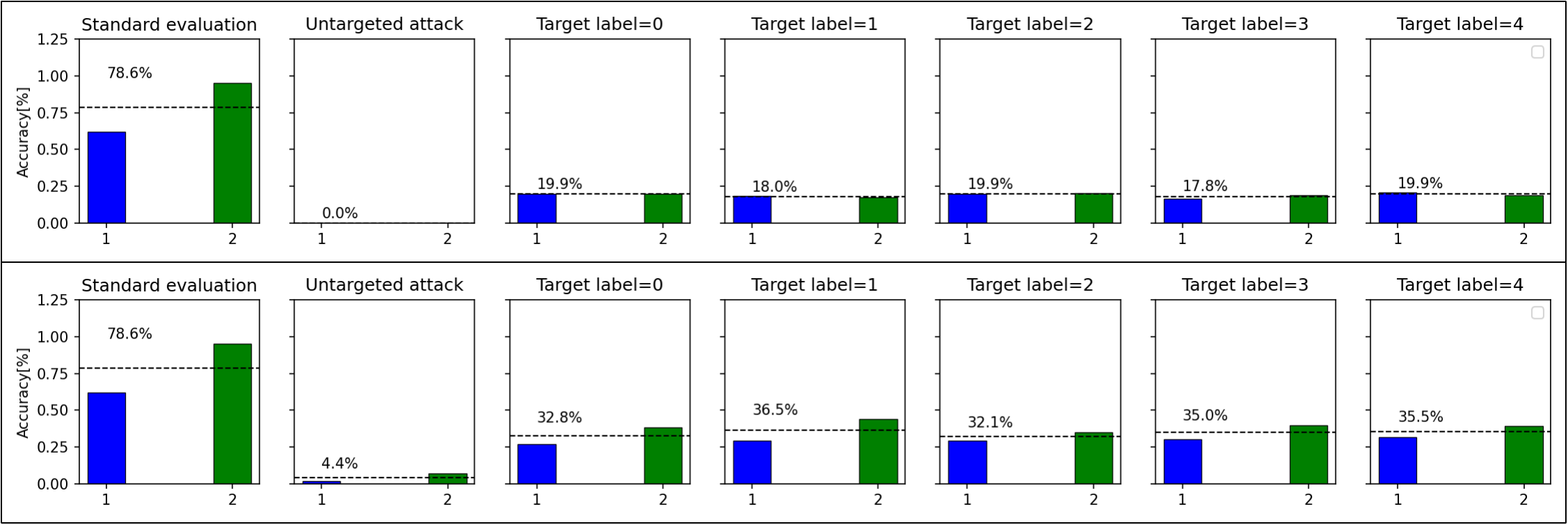}
\caption{Presents class-wise vulnerability of the EWC online \cite{ewc} against the FGSM and PGD \cite{fgsm,pgd} adversarial attacks under Domain-IL setting of continual learning.}
\label{fig:ewc_online_dil}
\end{figure}

In addition, Figure \ref{fig:ewc_online_dil} provides a visual representation of the class-wise vulnerability of the Elastic Weight Consolidation (EWC) online approach \cite{ewc} in the context of the Fast Gradient Sign Method (FGSM) adversarial attack \cite{fgsm}, specifically within the Domain-Incremental Learning (Domain-IL) setting, which is a paradigm of continual learning. The figure serves to illustrate the progressive impact of adversarial perturbations on the model's performance.

The first row of the figure displays the class-wise vulnerability of the EWC online approach to FGSM attacks, while the second row presents the class-wise vulnerability of another approach known as Incremental Class and Representation Learning (ICARL) against Projected Gradient Descent (PGD) attacks. Within each row, two sub-figures are showcased. The first sub-figure provides an overview of the average performance of the EWC online approach under standard evaluation conditions within the context of continual learning. In contrast, the second sub-figure within each row depicts the degradation in model performance resulting from untargeted adversarial attacks.

Subsequently, the subsequent sub-figures in both rows reveal the progressive deterioration of the average performance when subjected to targeted adversarial attacks. The headings associated with these sub-figures specify the labels that were the targets of these adversarial attacks, thereby highlighting the specific vulnerabilities of the model to certain classes. The x-axis in each sub-figure represents the task number, signifying the sequential order of tasks encountered during continual learning. Meanwhile, the y-axis quantifies the average accuracy of the model over ten independent runs, providing an indication of its overall performance. Notably, the horizontal bar in each sub-figure marks the average accuracy achieved across two distinct tasks, facilitating a comparative assessment of the model's stability and adaptability in the face of adversarial challenges.

\subsection{Analyzing Class-Wise Vulnerability of SI to Adversarial Attacks in Domain-IL Continual Learning}
Figure 1 illustrates the class-specific vulnerability of the Subject Instance (SI) model \cite{si} when subjected to Fast Gradient Sign Method (FGSM) adversarial attacks \cite{fgsm} within the context of Domain-Incremental Learning (Domain-IL). The figure is organized into two rows: the first row visualizes the class-wise vulnerability of SI against FGSM attacks, while the second row depicts its vulnerability to Projected Gradient Descent (PGD) attacks. Each row comprises sub-figures that convey different aspects of SI's performance under various evaluation conditions.
The initial sub-figure in both rows provides an overview of SI's average performance in a standard evaluation scenario for continual learning. Subsequently, the subsequent sub-plots within each row elucidate the extent of performance degradation experienced by SI under untargeted adversarial attacks. Furthermore, the subsequent sub-plots within each row showcase SI's decline in average performance when exposed to targeted adversarial attacks. The headers accompanying these sub-plots emphasize the specific target labels employed in the attacks.
To aid comprehension, the sub-plots x-axis represents the task number, while the y-axis denotes the average accuracy computed over ten independent runs. A horizontal bar on each sub-plot also illustrates the average accuracy achieved across two tasks. This visualization provides valuable insights into SI's resilience to adversarial attacks across different evaluation scenarios in the Domain-IL setting.

\begin{figure}[H]
\centering
\includegraphics[width=1.0\textwidth]{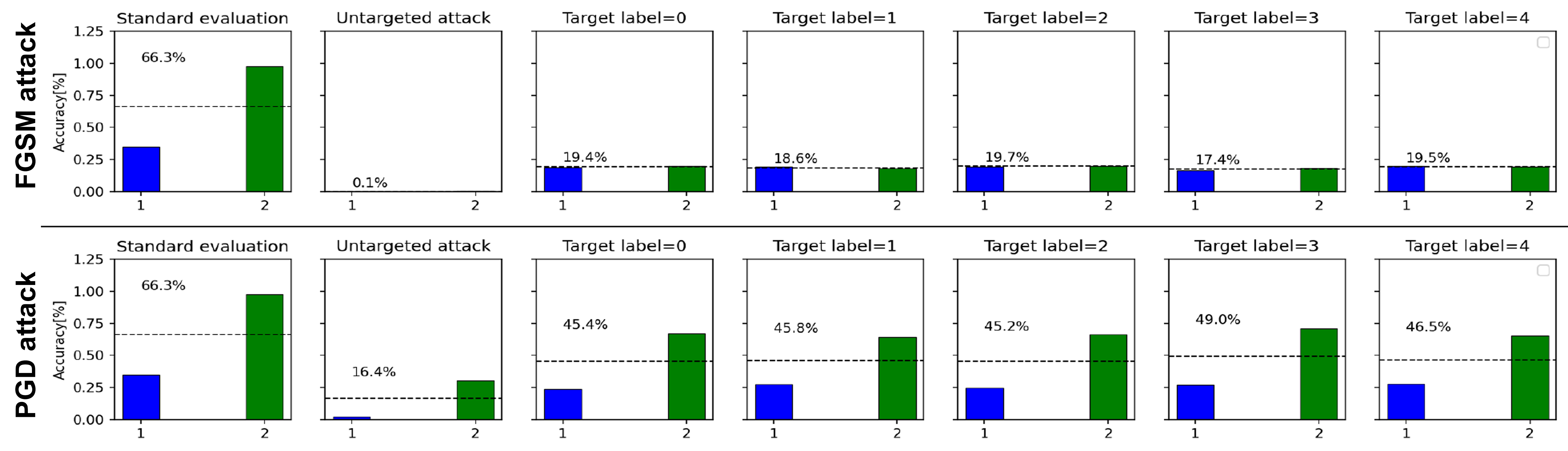}
\caption{Presents class-wise vulnerability of the SI \cite{si} against the FGSM and PGD \cite{fgsm,pgd} adversarial attacks under Domain-IL setting of continual learning.}
\label{fig:si_dil}
\end{figure}

\subsection{Analyzing Class-Wise Vulnerability of XDG to Adversarial Attacks in Domain-IL Continual Learning}
\begin{figure}[H]
\centering
\includegraphics[width=1.0\textwidth]{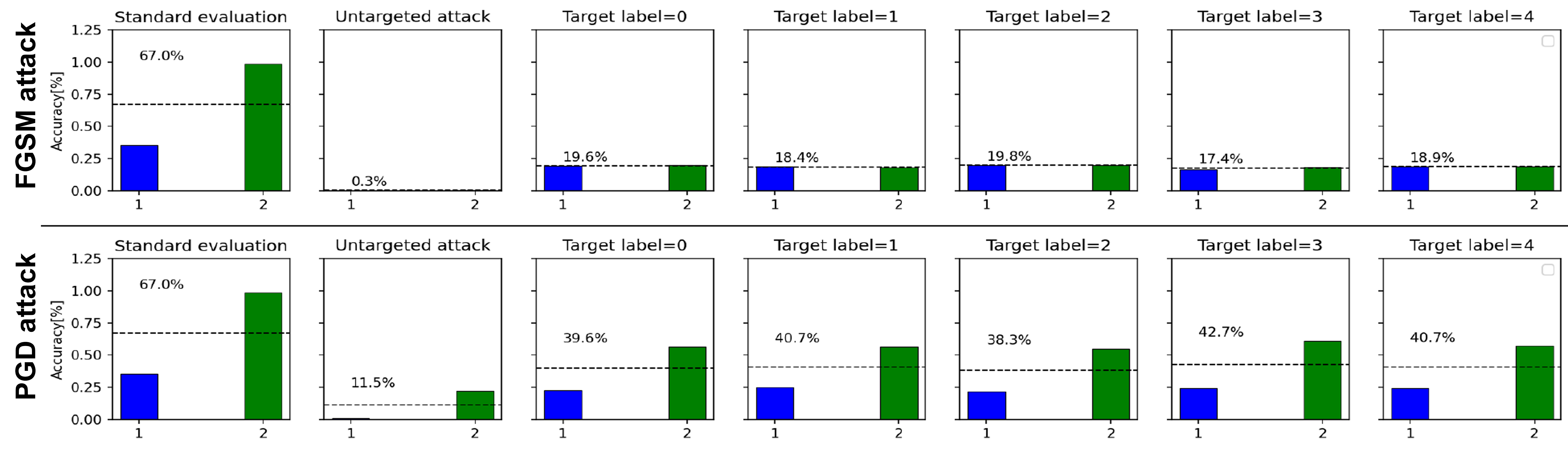}
\caption{Presents class-wise vulnerability of the XDG \cite{xgd} against the FGSM and PGD \cite{fgsm,pgd} adversarial attacks under Domain-IL setting of continual learning.}
\label{fig:xgd_dil}
\end{figure}

In a similar vein, Figure \ref{fig:xgd_dil} herein elucidates the class-wise susceptibility of the XDG model, as expounded in the work by [citation XGD], in the face of FGSM (Fast Gradient Sign Method) adversarial attacks within the context of Domain-Incremental Learning (Domain-IL). The presentation of this data is organized into two distinct rows, where the first row delineates the class-specific vulnerability of the XDG model to FGSM attacks, while the second row illustrates its susceptibility under PGD (Projected Gradient Descent) adversarial attacks.

Within each row, the graphical content consists of a set of sub-figures, serving as a comprehensive representation of the model's performance under various conditions. The initial sub-figure in both rows encapsulates the average performance of the XDG model, gauged through the lens of conventional continual learning evaluation protocols, devoid of adversarial perturbations. The subsequent sub-plots, positioned immediately below, succinctly depict the extent of performance degradation experienced by the XDG model in response to untargeted adversarial attacks, specifically FGSM in row 1 and PGD in row 2.

Subsequently, the sub-plots that ensue in each row divulge the model's deterioration in average performance in the wake of targeted adversarial attacks. Notably, these sub-plots are delineated by headers that provide a concise annotation of the specific labels being targeted in the respective adversarial attacks. The x-axis of each sub-plot is demarcated by task numbers, while the y-axis conveys the average accuracy achieved, which is derived from an aggregation of results across ten experimental runs. In an informative manner, the horizontal bar featured in each sub-plot signifies the average accuracy attained by the model over the course of two sequential tasks, thus encapsulating the model's performance across this span of tasks.

\subsection{Analyzing Class-Wise Vulnerability of DGR to Adversarial Attacks in Domain-IL Continual Learning}
\begin{figure}[H]
\centering
\includegraphics[width=1.0\textwidth]{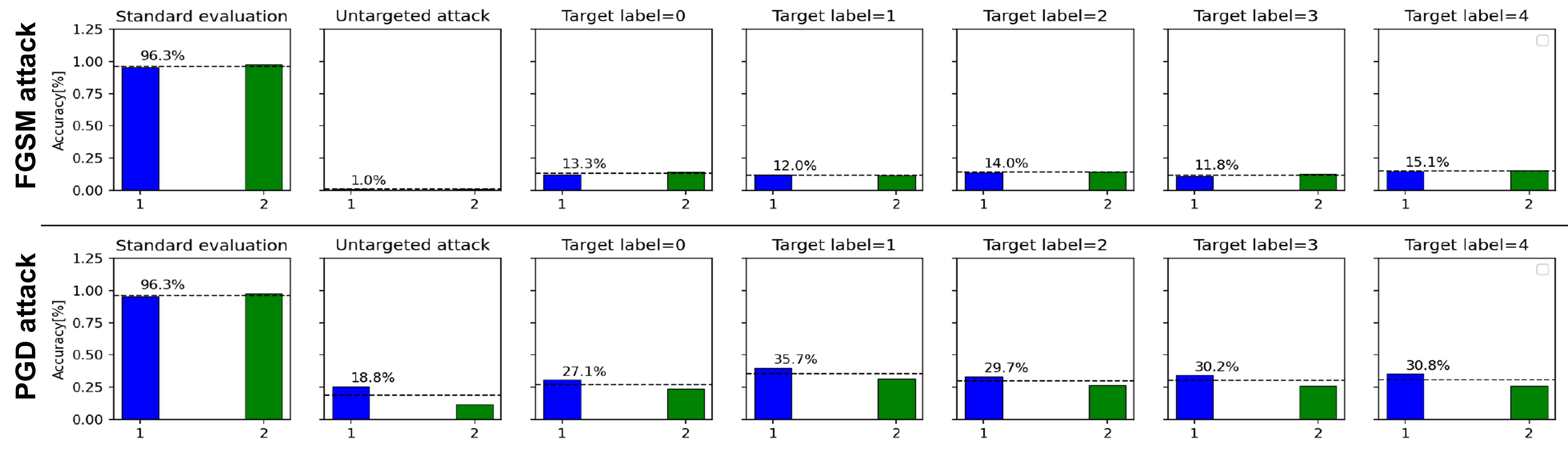}
\caption{Presents class-wise vulnerability of the DGR \cite{dgr} against the FGSM and PGD \cite{fgsm,pgd} adversarial attacks under Domain-IL setting of continual learning.}
\label{fig:dgr_dil}
\end{figure}

Figure \ref{fig:dgr_dil} illustrates the class-wise vulnerability of the Deep Generative Replay (DGR) model \cite{dgr} when subjected to Fast Gradient Sign Method (FGSM) adversarial attacks within the Domain-Incremental Learning (Domain-IL) paradigm. Specifically, the figure presents the DGR's class-wise vulnerability to FGSM attacks in row 1 and its susceptibility to Projected Gradient Descent (PGD) attacks in row 2. The first sub-figure showcases the DGR's average performance under standard evaluation conditions for continual learning in both rows. Subsequently, the second sub-plots in both rows depict the degradation in performance induced by untargeted adversarial attacks.
Furthermore, the subsequent sub-plots in both rows demonstrate how targeted adversarial attacks have further diminished the DGR's average performance. The headers of these sub-plots provide information about the targeted labels. Notably, the x-axis of the plots corresponds to the task number, while the y-axis represents the average accuracy calculated over ten independent runs. The horizontal bar in the plots represents the average accuracy across the two tasks.

\subsection{Analyzing Class-Wise Vulnerability of iCARL to Adversarial Attacks in Class-IL Continual Learning}
\begin{figure}[H]
\centering
\includegraphics[width=1.0\textwidth]{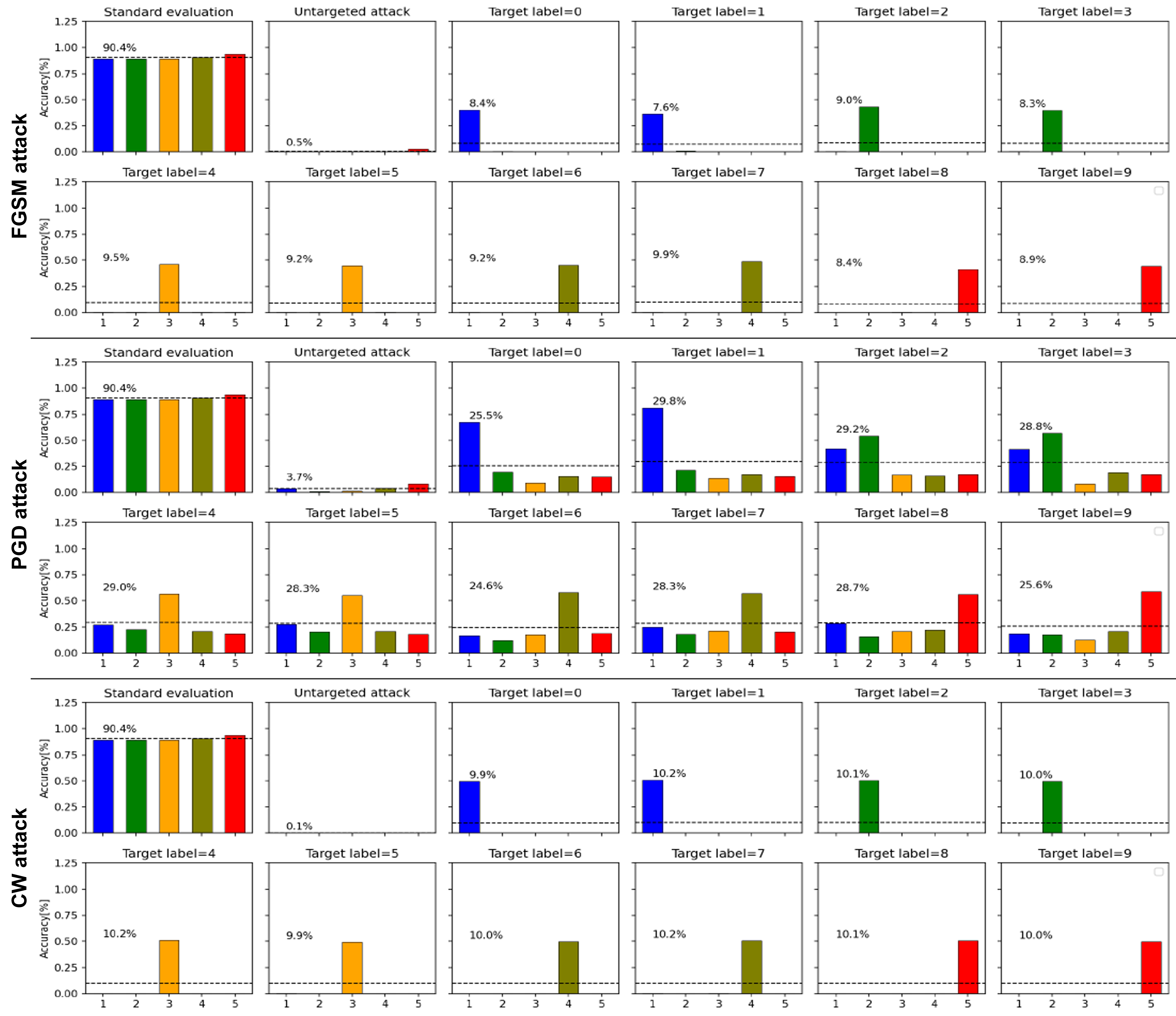}
\caption{Presents class-wise vulnerability of the ICARL \cite{icarl} against the FGSM, PGD, and CW \cite{fgsm,pgd,cw} adversarial attacks under Class-IL setting of continual learning.}
\label{fig:icarl_cil}
\end{figure}

Finally, Figure \ref{fig:icarl_cil} illustrates the class-wise vulnerability of the Incremental Classifier and Representation Learning (ICARL) model \cite{icarl} when subjected to adversarial attacks, including Fast Gradient Sign Method (FGSM), Projected Gradient Descent (PGD), and Carlini-Wagner (CW) attacks \cite{fgsm,pgd,cw}. These attacks are examined within the Class-Incremental Learning (Class-IL) framework, a subdomain of continual learning.
The figure is organized into six rows, each focusing on a specific type of attack. The first and second rows showcase the ICARL's class-wise vulnerability against FGSM attacks. Similarly, the third and fourth rows present the ICARL's susceptibility to PGD attacks, and the fifth and sixth rows depict the ICARL's vulnerability to CW attacks.
Within each row, the first sub-figure illustrates the ICARL's performance, measured in terms of the Lwf (Learning without Forgetting) metric, under standard evaluation conditions for continual learning. The second sub-plot in rows 1 and 2 provides insight into the degradation in performance resulting from untargeted adversarial attacks. The headers of these sub-plots specify the labels that are the targets of the adversarial attacks, indicating their specific focus.
Furthermore, the x-axis in the plots corresponds to the task number, while the y-axis represents the average accuracy calculated over ten independent runs. The horizontal bar in the plots signifies the average accuracy across all tasks considered in the evaluation.

The individual class-level vulnerabilities of the continual learning algorithms within the Domain-Incremental Learning (Domain-IL) context are depicted in Figures \ref{fig:ewc_dil}, \ref{fig:ewc_online_dil}, \ref{fig:si_dil}, \ref{fig:xgd_dil}, and \ref{fig:dgr_dil}. Similarly, Figures \ref{fig:icarl_cil} showcase the individual class-level vulnerability of these algorithms in the context of Class-Incremental Learning (Class-IL).

\subsection{Adversarial Attack in Task-IL settings}

Figures \ref{fig:ewc_til}, \ref{fig:ewc_online_til}, \ref{fig:si_til}, \ref{fig:lwf_til}, \ref{fig:dgr_til}, and \ref{fig:dgr_distill_til} serve as visual representations of the class-specific and average accuracy reductions observed when employing EWC, EWC online, SI, Lwf, DGR, and DGR+Distill methods in the face of FGSM, PGD, and CW adversarial attacks, respectively.

\subsection{Adversarial Attack in Domain-IL settings}

The figures designated as Figure \ref{fig:ewc_dil}, Figure \ref{fig:ewc_online_dil}, Figure \ref{fig:si_dil}, Figure \ref{fig:xgd_dil}, and Figure \ref{fig:dgr_dil} serve to depict the class-specific and aggregate decrements in accuracy observed within the context of domain incremental training, with respect to the following adversarial attack methods: Fast Gradient Sign Method (FGSM) and Projected Gradient Descent (PGD). These figures provide a visual representation of the performance of various techniques, namely Elastic Weight Consolidation (EWC), Online EWC, Synaptic Intelligence (SI), XDG, and Deep Generative Replay (DGR), in the face of such adversarial challenges.

\subsection{Adversarial Attack in Class-IL settings} 
Figure \ref{fig:icarl_cil} depicts the class-wise and average declines in accuracy exhibited by the ICARL method when subjected to FGSM, PGD, and CW adversarial attacks in a class incremental training setting. Our empirical findings underscore the imperative to embark upon a novel research trajectory akin to the domains of robust machine learning and trustworthy machine learning, with a central emphasis on crafting methodologies that not only mitigate catastrophic forgetting but also proffer robust guarantees against any potential compromise of the security pertaining to previously acquired knowledge. Frail defensive mechanisms against adversarial examples substantially curtail the practical utility of continual learning-based solutions, undermining the fundamental objectives of continual learning algorithms and artificial agents.

For example, the vulnerability of a continual learning system to slight modifications of a malware file, resulting in its erroneous classification as benign within an older or current task, renders the malware classifier ineffectual. It is imperative for continual learning algorithms to counteract catastrophic forgetting and demonstrate robust resilience, rendering them deploy able in security-critical real-world environments. Our findings underscore the susceptibility of virtually all state-of-the-art continual learning methods to severe vulnerabilities posed by adversarial attacks. In light of these observations, we advocate for incorporating additional metrics by the continual learning research community, which extend beyond evaluating models solely with respect to mitigating catastrophic forgetting to encompass the assessment of methodological robustness and its implications for model security.

In conclusion, our observations reveal that the susceptibility of algorithms to generate false memories is exacerbated by the ease with which learning activities can be misclassified. Furthermore, our findings indicate that the propensity of algorithms to produce false memories stems from the susceptibility of learning activities to incorrect classification.

\section{Conclusion}
We have substantiated that the capacity to misclassify any class, whether belonging to current or previously acquired tasks, can be readily exploited by creating adversarial examples targeting a specific desired class. Our investigations reveal the effectiveness of adversarial attacks across three distinct scenarios: task-incremental learning, domain-incremental learning, and class-incremental learning. These scenarios encompass evaluations of various state-of-the-art continual learning methodologies, including EWC, EWC online, SI, XDG, LwF, DGR, DGR+Distill, and iCARL.

Our empirical findings strongly underscore the heightened vulnerability of continual learning techniques to adversarial attacks. In light of these vulnerabilities, we emphatically advocate for a concerted effort within the research community to bolster the robustness of continuous learning paradigms. It is imperative to address these vulnerabilities comprehensively to prevent such detrimental situations from materializing in practical applications.

\section{Future Research Directions}
We propose the incorporation of adversarial training into the training phase of each task, in conjunction with standard supervised learning, as a means to enhance our understanding of the robustness of individual models. Our objective is to equip continual learning models with the capability to withstand adversarial attacks and mitigate the risk of false memory formation in tasks learned over time. It is worth noting that other potentially confounding factors, such as acute issues like Backdoor attacks on previously acquired tasks, could provide valuable insights into the resilience of earlier tasks when subjected to these approaches. Consequently, we advocate for further theoretical research to elucidate the role of these factors in enhancing the security of continual learning. Notably, our empirical findings reveal that continual learning techniques exhibit heightened vulnerability to adversarial attacks.

\bibliographystyle{unsrt}
\bibliography{main.bib}

\begin{thebibliography}{10}

\bibitem{cai2020review}
Lei Cai, Jingyang Gao, and Di~Zhao.
\newblock A review of the application of deep learning in medical image
  classification and segmentation.
\newblock {\em Annals of translational medicine}, 8(11), 2020.

\bibitem{otter2020survey}
Daniel~W Otter, Julian~R Medina, and Jugal~K Kalita.
\newblock A survey of the usages of deep learning for natural language
  processing.
\newblock {\em IEEE transactions on neural networks and learning systems},
  32(2):604--624, 2020.

\bibitem{jeong2022systematic}
Jiwoong~J Jeong, Amara Tariq, Tobiloba Adejumo, Hari Trivedi, Judy~W Gichoya,
  and Imon Banerjee.
\newblock Systematic review of generative adversarial networks (gans) for
  medical image classification and segmentation.
\newblock {\em Journal of Digital Imaging}, pages 1--16, 2022.

\bibitem{pandey2021comprehensive}
Babita Pandey, Devendra~Kumar Pandey, Brijendra~Pratap Mishra, and Wasiur
  Rhmann.
\newblock A comprehensive survey of deep learning in the field of medical
  imaging and medical natural language processing: Challenges and research
  directions.
\newblock {\em Journal of King Saud University-Computer and Information
  Sciences}, 2021.

\bibitem{yadav2020sentiment}
Ashima Yadav and Dinesh~Kumar Vishwakarma.
\newblock Sentiment analysis using deep learning architectures: a review.
\newblock {\em Artificial Intelligence Review}, 53(6):4335--4385, 2020.

\bibitem{le2021deep}
Ngan Le, Vidhiwar~Singh Rathour, Kashu Yamazaki, Khoa Luu, and Marios Savvides.
\newblock Deep reinforcement learning in computer vision: a comprehensive
  survey.
\newblock {\em Artificial Intelligence Review}, pages 1--87, 2021.

\bibitem{khan2020cascading}
Khan Hikmat, Pir~Masoom Shah, Munam~Ali Shah, Saif ul~Islam, and Joel~JPC
  Rodrigues.
\newblock Cascading handcrafted features and convolutional neural network for
  iot-enabled brain tumor segmentation.
\newblock {\em Computer Communications}, 153:196--207, 2020.

\bibitem{Hikmat_75}
Khan Hikmat, Rasool Ghulam, Bouaynaya Nidhal, C, and Johnson Charles~C.
\newblock Rotorcraft flight information inference from cockpit videos using
  deep learning.
\newblock {\em American Helicopter Society 75th Annual Forum, Philadelphia,
  Pennsylvania, USA}, May 2019.

\bibitem{Hikmat_76}
Khan Hikmat, Rasool Ghulam, Bouaynaya Nidhal, C, , Travis Tyler, Thompson
  Lacey, and Johnson Charles~C.
\newblock Explainable ai: Rotorcraft attitude prediction.
\newblock {\em Vertical Flight Society’s 76th Annual Forum and Technology
  Display, Virginia Beach, Virginia, USA}, Oct 2020.

\bibitem{Hikmat_77}
Khan Hikmat, Rasool Ghulam, Bouaynaya Nidhal, C, , Travis Tyler, Thompson
  Lacey, and Johnson Charles~C.
\newblock Deep ensemble for rotorcraft attitude prediction.
\newblock {\em Vertical Flight Society’s 77th Annual Forum and Technology
  Display, Palm Beach, Florida, USA}, May 2021.

\bibitem{zeng2022introduction}
Xiangming Zeng and Liangqu Long.
\newblock Introduction to artificial intelligence.
\newblock In {\em Beginning Deep Learning with TensorFlow}, pages 1--45.
  Springer, 2022.

\bibitem{parisi2019continual}
German~I Parisi, Ronald Kemker, Jose~L Part, Christopher Kanan, and Stefan
  Wermter.
\newblock Continual lifelong learning with neural networks: A review.
\newblock {\em Neural Networks}, 113:54--71, 2019.

\bibitem{delange2021continual}
Matthias Delange, Rahaf Aljundi, Marc Masana, Sarah Parisot, Xu~Jia, Ales
  Leonardis, Greg Slabaugh, and Tinne Tuytelaars.
\newblock A continual learning survey: Defying forgetting in classification
  tasks.
\newblock {\em IEEE Transactions on Pattern Analysis and Machine Intelligence},
  2021.

\bibitem{wang2019security}
Xianmin Wang, Jing Li, Xiaohui Kuang, Yu-an Tan, and Jin Li.
\newblock The security of machine learning in an adversarial setting: A survey.
\newblock {\em Journal of Parallel and Distributed Computing}, 130:12--23,
  2019.

\bibitem{chakraborty2018adversarial}
Anirban Chakraborty, Manaar Alam, Vishal Dey, Anupam Chattopadhyay, and Debdeep
  Mukhopadhyay.
\newblock Adversarial attacks and defences: A survey.
\newblock {\em arXiv preprint arXiv:1810.00069}, 2018.

\bibitem{ewc}
James Kirkpatrick, Razvan Pascanu, Neil Rabinowitz, Joel Veness, Guillaume
  Desjardins, Andrei~A Rusu, Kieran Milan, John Quan, Tiago Ramalho, Agnieszka
  Grabska-Barwinska, et~al.
\newblock Overcoming catastrophic forgetting in neural networks.
\newblock {\em Proceedings of the national academy of sciences},
  114(13):3521--3526, 2017.

\bibitem{si}
Friedemann Zenke, Ben Poole, and Surya Ganguli.
\newblock Continual learning through synaptic intelligence.
\newblock {\em Proceedings of machine learning research}, 70:3987, 2017.

\bibitem{lwf}
Zhizhong Li and Derek Hoiem.
\newblock Learning without forgetting.
\newblock {\em IEEE transactions on pattern analysis and machine intelligence},
  40(12):2935--2947, 2017.

\bibitem{dgr}
Hanul Shin, Jung~Kwon Lee, Jaehong Kim, and Jiwon Kim.
\newblock Continual learning with deep generative replay.
\newblock In {\em Advances in Neural Information Processing Systems}, pages
  2990--2999, 2017.

\bibitem{ring1994continual}
Mark~Bishop Ring.
\newblock Continual learning in reinforcement environments.
\newblock {\em PhD thesis, University of Texas at Austin}, 1994.

\bibitem{thrun1995lifelong}
Sebastian Thrun and Tom~M Mitchell.
\newblock Lifelong robot learning.
\newblock {\em Robotics and autonomous systems}, 15(1-2):25--46, 1995.

\bibitem{mccloskey1989catastrophic}
Michael McCloskey and Neal~J Cohen.
\newblock Catastrophic interference in connectionist networks: The sequential
  learning problem.
\newblock In {\em Psychology of learning and motivation}, volume~24, pages
  109--165. Elsevier, 1989.

\bibitem{ratcliff1990connectionist}
Roger Ratcliff.
\newblock Connectionist models of recognition memory: constraints imposed by
  learning and forgetting functions.
\newblock {\em Psychological review}, 97(2):285, 1990.

\bibitem{qu2021recent}
Haoxuan Qu, Hossein Rahmani, Li~Xu, Bryan Williams, and Jun Liu.
\newblock Recent advances of continual learning in computer vision: An
  overview.
\newblock {\em arXiv preprint arXiv:2109.11369}, 2021.

\bibitem{li2017learning}
Zhizhong Li and Derek Hoiem.
\newblock Learning without forgetting.
\newblock {\em IEEE transactions on pattern analysis and machine intelligence},
  40(12):2935--2947, 2017.

\bibitem{dhar2019learning}
Prithviraj Dhar, Rajat~Vikram Singh, Kuan-Chuan Peng, Ziyan Wu, and Rama
  Chellappa.
\newblock Learning without memorizing.
\newblock In {\em Proceedings of the IEEE/CVF Conference on Computer Vision and
  Pattern Recognition}, pages 5138--5146, 2019.

\bibitem{kirkpatrick2017overcoming}
James Kirkpatrick, Razvan Pascanu, Neil Rabinowitz, Joel Veness, Guillaume
  Desjardins, Andrei~A Rusu, Kieran Milan, John Quan, Tiago Ramalho, Agnieszka
  Grabska-Barwinska, et~al.
\newblock Overcoming catastrophic forgetting in neural networks.
\newblock {\em Proceedings of the National Academy of Sciences of the United
  States of America}, 114(13):3521--3526, 2017.

\bibitem{zenke2017continual}
F~Zenke, B~Poole, and S~Ganguli.
\newblock Continual learning through synaptic intelligence.
\newblock {\em Proceedings of machine learning research}, 70:3987--3995, 2017.

\bibitem{aljundi2018memory}
Rahaf Aljundi, Francesca Babiloni, Mohamed Elhoseiny, Marcus Rohrbach, and
  Tinne Tuytelaars.
\newblock Memory aware synapses: Learning what (not) to forget.
\newblock In {\em Proceedings of the European Conference on Computer Vision
  (ECCV)}, pages 139--154, 2018.

\bibitem{saha2020gradient}
Gobinda Saha, Isha Garg, and Kaushik Roy.
\newblock Gradient projection memory for continual learning.
\newblock In {\em International Conference on Learning Representations (ICLR)},
  2020.

\bibitem{Ergn2020ContinualLW}
Esra Erg{\"u}n and Behçet~Uğur T{\"o}reyin.
\newblock Continual learning with sparse progressive neural networks.
\newblock {\em 2020 28th Signal Processing and Communications Applications
  Conference (SIU)}, pages 1--4, 2020.

\bibitem{rusu2016progressive}
Andrei~A Rusu, Neil~C Rabinowitz, Guillaume Desjardins, Hubert Soyer, James
  Kirkpatrick, Koray Kavukcuoglu, Razvan Pascanu, and Raia Hadsell.
\newblock Progressive neural networks.
\newblock {\em arXiv preprint arXiv:1606.04671}, 2016.

\bibitem{icarl}
Sylvestre-Alvise Rebuffi, Alexander Kolesnikov, Georg Sperl, and Christoph~H
  Lampert.
\newblock icarl: Incremental classifier and representation learning.
\newblock In {\em Proceedings of the IEEE conference on Computer Vision and
  Pattern Recognition}, pages 2001--2010, 2017.

\bibitem{lopez2017gradient}
David Lopez-Paz and Marc'Aurelio Ranzato.
\newblock Gradient episodic memory for continual learning.
\newblock {\em Advances in Neural Information Processing Systems (NIPS)}, 30,
  2017.

\bibitem{robins1995catastrophic}
Anthony Robins.
\newblock Catastrophic forgetting, rehearsal and pseudorehearsal.
\newblock {\em Connection Science}, 7(2):123--146, 1995.

\bibitem{riemer2018learning}
Matthew Riemer, Ignacio Cases, Robert Ajemian, Miao Liu, Irina Rish, Yuhai Tu,
  , and Gerald Tesauro.
\newblock Learning to learn without forgetting by maximizing transfer and
  minimizing interference.
\newblock In {\em International Conference on Learning Representations (ICLR)},
  2019.

\bibitem{chaudhry2020continual}
Arslan Chaudhry, Naeemullah Khan, Puneet Dokania, and Philip Torr.
\newblock Continual learning in low-rank orthogonal subspaces.
\newblock {\em Advances in Neural Information Processing Systems (NIPS)},
  33:9900--9911, 2020.

\bibitem{fgsm}
Ian~J Goodfellow, Jonathon Shlens, and Christian Szegedy.
\newblock Explaining and harnessing adversarial examples.
\newblock {\em arXiv preprint arXiv:1412.6572}, 2014.

\bibitem{pgd}
Aleksander Madry, Aleksandar Makelov, Ludwig Schmidt, Dimitris Tsipras, and
  Adrian Vladu.
\newblock Towards deep learning models resistant to adversarial attacks.
\newblock {\em arXiv preprint arXiv:1706.06083}, 2017.

\bibitem{cw}
Nicholas Carlini and David Wagner.
\newblock Towards evaluating the robustness of neural networks.
\newblock In {\em 2017 ieee symposium on security and privacy (sp)}, pages
  39--57. IEEE, 2017.

\bibitem{three_scenarios}
Gido~M van~de Ven and Andreas~S Tolias.
\newblock Three scenarios for continual learning.
\newblock {\em arXiv}, pages arXiv--1904, 2019.

\bibitem{lecun2010mnist}
Yann LeCun, Corinna Cortes, and CJ~Burges.
\newblock Mnist handwritten digit database.
\newblock 2010.

\bibitem{xgd}
Nicolas~Y Masse, Gregory~D Grant, and David~J Freedman.
\newblock Alleviating catastrophic forgetting using context-dependent gating
  and synaptic stabilization.
\newblock {\em Proceedings of the National Academy of Sciences},
  115(44):E10467--E10475, 2018.

\bibitem{rauber2017foolboxnative}
Jonas Rauber, Roland Zimmermann, Matthias Bethge, and Wieland Brendel.
\newblock Foolbox native: Fast adversarial attacks to benchmark the robustness
  of machine learning models in pytorch, tensorflow, and jax.
\newblock {\em Journal of Open Source Software}, 5(53):2607, 2020.

\end{thebibliography}
\end{document}